\documentclass[10pt,twocolumn,letterpaper]{article}

\usepackage[pagenumbers]{cvpr} % To force page numbers, e.g. for an arXiv version
\usepackage{algorithm}
\usepackage{algorithmic}
\usepackage{multirow}
\usepackage{makecell}
\usepackage{pifont}
\definecolor{cvprblue}{rgb}{0.21,0.49,0.74}
\usepackage[pagebackref,breaklinks,colorlinks,allcolors=cvprblue]{hyperref}

\def\paperID{30523} % *** Enter the Paper ID here
\def\confName{CVPR}
\def\confYear{2026}

\title{Video Signature: Implicit Watermarking for Video Diffusion Models}

\author{
  Yu Huang$^{1}$\thanks{Equal Contribution: yhuang489@connect.hkust-gz.edu.cn},~
  Junhao Chen$^{1}$\footnotemark[1],~
  Shuliang Liu$^{1}$, ~
  Hanqian Li$^{1}$, ~\\
  Jungang Li$^{1}$,~
  Qi Zheng$^{1}$,~
  Aiwei Liu$^{3}$,~
  Yi R. (May) Fung$^{2}$, ~ 
  Xuming Hu$^{1, 2}$\thanks{Corresponding author: xuminghu@hkust-gz.edu.cn.}
  \\
  \\
  $^{1}$AI Thrust, Hong Kong University of Science and Technology (Guangzhou), China \\
  $^{2}$Hong Kong University of Science and Technology, Hong Kong, China \\
  $^{3}$School of Software, BNRist, Tsinghua University, China
}

\begin{document}
\maketitle
\begin{abstract}

The rapid development of Artificial Intelligence Generated Content (AIGC) has led to significant progress in video generation, but also raises serious concerns about intellectual property protection and reliable content tracing. Watermarking is a widely adopted solution to this issue, yet existing methods for video generation mainly follow a post-generation paradigm, which often fails to effectively balance the trade-off between video quality and watermark extraction. Meanwhile, current in-generation methods that embed the watermark into the initial Gaussian noise usually incur substantial additional computation. To address these issues, we propose \textbf{Video Signature} (\textsc{VidSig}), an implicit watermarking method for video diffusion models that enables imperceptible and adaptive watermark integration during video generation with almost no extra latency. Specifically, we partially fine-tune the latent decoder, where \textbf{Perturbation-Aware Suppression} (PAS) pre-identifies and freezes perceptually sensitive layers to preserve visual quality. Beyond spatial fidelity, we further enhance temporal consistency by introducing a lightweight \textbf{Temporal Alignment} module that guides the decoder to generate coherent frame sequences during fine-tuning. Experimental results show that \textsc{VidSig} achieves the best trade-off among watermark extraction accuracy, video quality, and watermark latency. It also demonstrates strong robustness against both spatial and temporal tamper, and remains stable across different video lengths and resolutions, highlighting its practicality in real-world scenarios.

\end{abstract}

%% Template
% \begin{abstract}
% The ABSTRACT is to be in fully justified italicized text, at the top of the left-hand column, below the author and affiliation information.
% Use the word ``Abstract'' as the title, in 12-point Times, boldface type, centered relative to the column, initially capitalized.
% The abstract is to be in 10-point, single-spaced type.
% Leave two blank lines after the Abstract, then begin the main text.
% Look at previous \confName abstracts to get a feel for style and length.
% \end{abstract}    
\section{Introduction}
\label{section: Introduction}
With the rapid advancement of Artificial Intelligence Generated Content (AIGC), remarkable progress has been achieved across various generative modalities, including texts~\citep{achiam2023gpt, touvron2023llama, yang2025qwen3}, images~\citep{ho2020denoising, rombach2022high, peebles2023scalable}, audios~\citep{huang2024audiogpt, zhang2023speechgpt, xu2025qwen2}, videos~\citep{blattmann2023stable, hong2022cogvideo, kong2024hunyuanvideo,liu2025javisgpt,xun2025rtv} and unified generation~\citep{liu2025javisgpt,cui2025emu3}. However, unauthorized use or misuse of these models can lead to significant risks, such as deepfake and some ethical concerns~\citep{westerlund2019emergence, zohny2023ethics}. Among the various techniques proposed to address this critical issue, watermarking technology serves as a promising solution for asserting ownership and enabling provenance tracking of the generated content\citep{liang2024watermarking, liu2024survey, hu2025videomark, liu2025vla, su2025safe}. However, despite the rapid progress of watermarking in text~\citep{kirchenbauer2023watermark, liu2025vla} and image domains~\citep{wen2023tree, yang2024gaussian}, its extension to videos remains relatively underexplored and presents unique challenges.

\begin{figure}[t]
    \centering
    \includegraphics[width=0.95\linewidth]{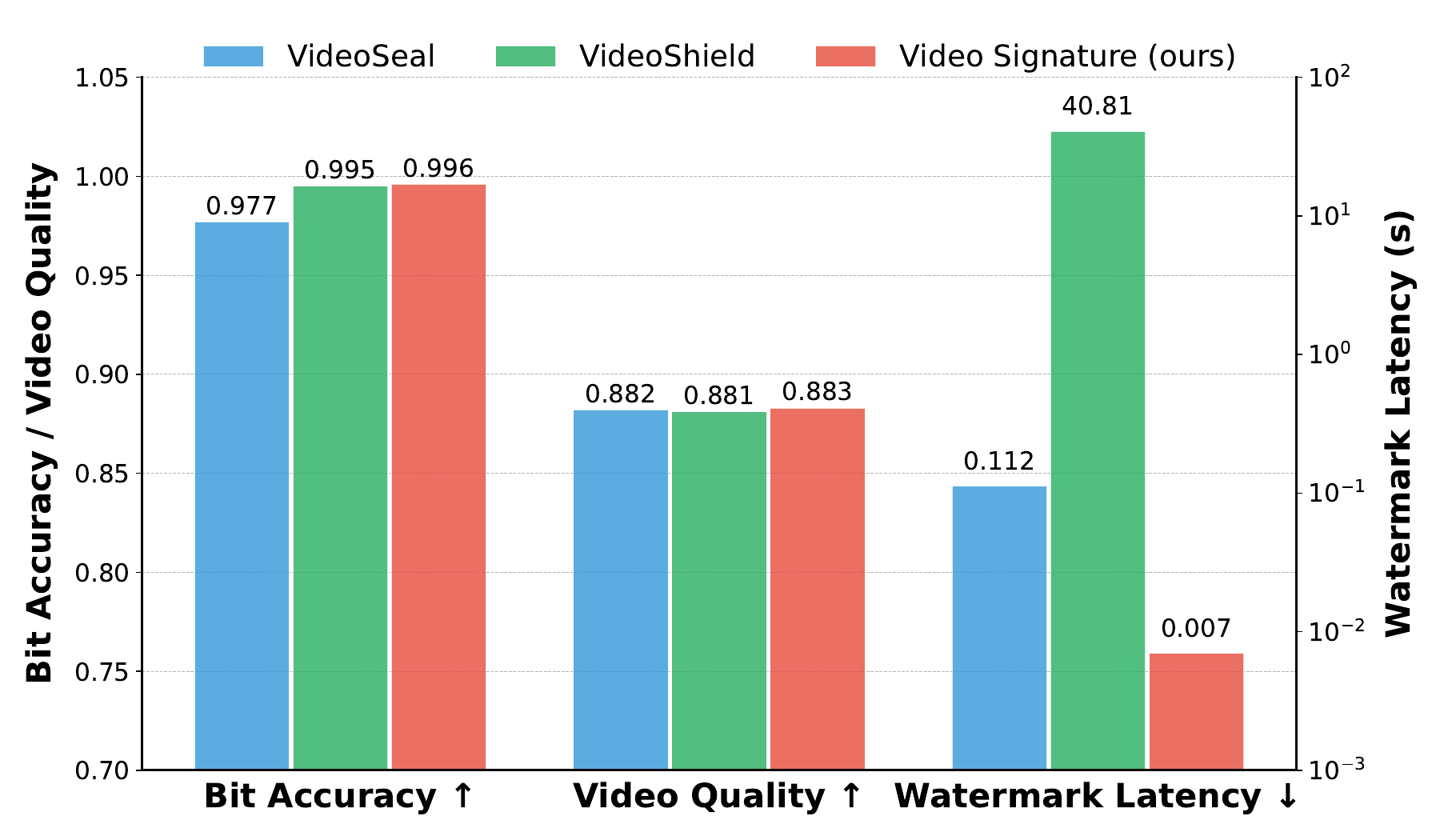}
    \vspace{-5pt}
    \caption{Comparison between Video Signature and other watermarking methods in watermark extraction accuracy, video quality (measured by VBench~\citep{huang2024vbench}) and watermark latency. The watermark latency is the summation of the latency for watermark embedding and extraction. The values of the metrics are the average of T2V model and I2V model, detailed information see Section~\ref{section: EXP}.}
    \vspace{-15pt}
    \label{fig:introduction}
\end{figure}

Many current watermarking methods for video generation follow a post-generation paradigm that separates the generation and watermarking processes. The watermark is embedded into the video after generation through some robust algorithms, from early traditional methods such as Spread Spectrum~\citep{cox1996secure}, DCT~\citep{barni1998dct}, and DWT~\citep{kundur1998digital} to current deep neural networks, including RivaGAN~\citep{zhang2019robust} and VideoSeal~\citep{fernandez2024video}. This paradigm not only introduces extra computational overhead but also often fails to strike a balance between visual quality and watermark extraction accuracy, making it less reliable and effective in practice. Recent efforts have explored in-generation methods, which embed watermarks during the generative process. As a pioneer, VideoShield~\citep{hu2025videoshield} extends Gaussian shading~\citep{yang2024gaussian} from the image domain to the video domain by embedding a multi-bit watermark into the initial Gaussian noise, allowing the watermark to be injected into every generated frame. VideoMark~\citep{hu2025videomark} further leverages pseudorandom error correction code to embed much longer multi-bit messages. These methods then leverage DDIMs~\citep{song2020denoising} to recover an approximate initial Gaussian noise, subsequently detecting the watermark on it. However, despite the strong extraction accuracy and visual quality preservation, these methods suffer from the extremely high computational cost for watermark embedding and extraction. Another line of in-generation methods in the image domain fine-tunes the latent decoder to embed the watermark during the mapping from latent space to image space~\citep{fernandez2023stable, rezaei2024lawa, kim2024wouaf}, and extracts the watermark using an external extractor aligned with the fine-tuning process. These methods enable watermark insertion with negligible latency. However, directly applying such methods to video generation overlooks the temporal consistency of video content, and exhaustively fine-tuning the entire latent decoder often leads to noticeable visual artifacts.

To address the aforementioned problems, we propose \textbf{Video Signature} (\textsc{VidSig}), a framework that integrates the watermark into \textit{each frame} of the generated video without additional modification to the model architecture and initial Gaussian noise. We embed the watermark message into each frame due to the vulnerability of video data to temporal attacks, such as frame dropping or shuffling. Specifically, we follow the pipeline of the training-based method to fine-tune the latent decoder to embed invisible watermarks into the video during generation. Before fine-tuning, we adopt a \textbf{Perturbation-Aware Suppression (PAS)} search algorithm to pre-identify and freeze perceptually sensitive layers to preserve visual quality. Furthermore, to capture temporal consistency, we introduce a straightforward but effective \textbf{Temporal Alignment} module that guides the decoder to produce coherent frame sequences during fine-tuning. The experiments on text-to-video (T2V) and image-to-video (I2V) models shown in Figure~\ref{fig:introduction} demonstrate that \textsc{VidSig} achieves the best trade-off among watermark extraction accuracy, video quality, and watermark latency. It attains a near-perfect bit accuracy of 0.996 while preserving video quality comparable to other methods, and it also introduces the lowest watermark latency. In summary, our main contributions are as follows:
\begin{itemize}
    \item[\ding{182}] We highlight the challenges of watermarking in video generative models and propose \textbf{Video Signature} (\textsc{VidSig}), an implicit watermarking method that embeds watermark messages directly into the video generation process by fine-tuning the latent decoder.
    \item[\ding{183}] We introduce Perturbation-Aware Suppression (PAS), a search algorithm to efficiently identify perceptually sensitive layers, and we introduce a Temporal Alignment module that enforces the inter-frame coherence.
    \item[\ding{184}] Experimental results show that \textsc{VidSig} achieves the best overall performance in watermark extraction, visual quality, and generation efficiency. It also exhibits strong robustness against both spatial and temporal tampering, and remains stable across different video lengths and resolutions, highlighting its practicality in real-world scenarios.
\end{itemize}

\section{Related Work}

\begin{figure*}[t]
    \centering
    \includegraphics[width=0.95\textwidth]{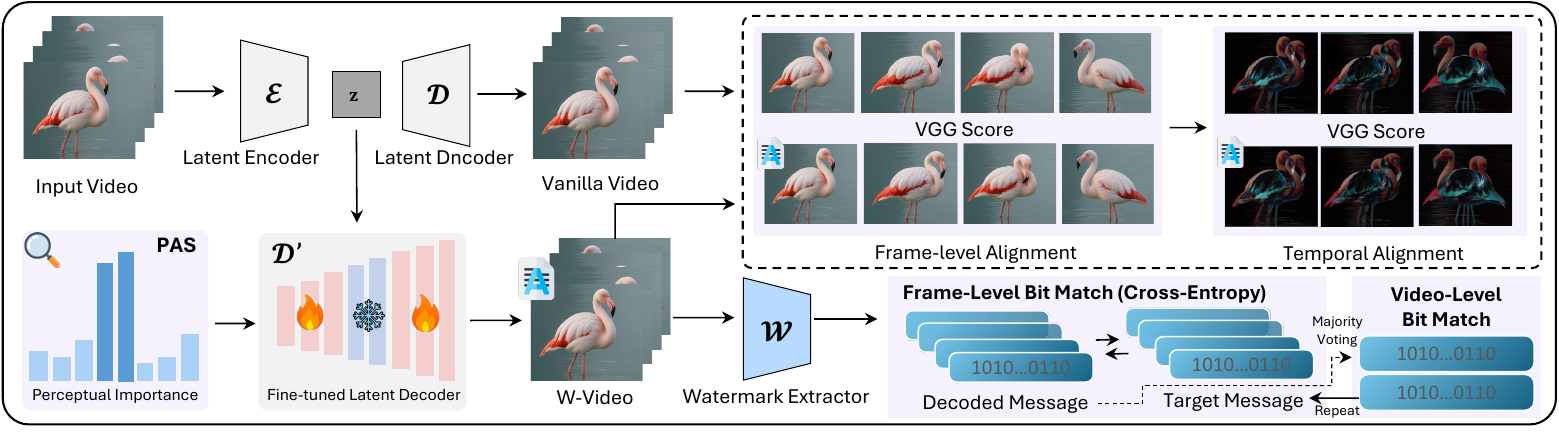}
    \caption{The training pipeline of Video Signature. \textbf{(1)} Given an input video, we first encode it into a latent representation and decode it with a frozen latent decoder. \textbf{(2)} Before optimization, the proposed PAS module searches the most perceptually sensitive layers and freezes them. \textbf{(3)} The watermarked decoder \(\mathcal{D}\)' is then optimized to embed a secret key into the generated video with three different objectives: pixel-level alignment, inter-frame level alignment, and bit match.}
    \label{fig:pipeline}
\end{figure*}

\paragraph{Diffusion-Based Video Generation}

Recent video generation models are mainly built on Latent Diffusion Models (LDMs)~\citep{rombach2022high}, ModelScope~\citep{wang2023modelscope}, which uses a 2D VAE to compress each frame into the latent space, employs a U-Net based architecture for denoising across both spatial and temporal dimensions. Stable Video Diffusion~\citep{blattmann2023stable}, which is also built on U-Net, uses a 3D VAE to encode the entire video into the latent space, allowing the models to naturally capture spatial and temporal relationships simultaneously. Models like Latte~\citep{ma2024latte}, Open-sora~\citep{zheng2024open}, ViDu~\citep{bao2024vidu}, uses the more scalable DiT-based~\citep{peebles2023scalable} architecture to learn the denoising process. The release of Sora by OpenAI~\citep{brooks2024video} and Hunyuan Video by Tencent~\citep{kong2024hunyuanvideo} has enabled the generation of longer and higher-quality videos, but also raises critical concerns about the misuse of such models~\citep{westerlund2019emergence, zohny2023ethics}.

\vspace{-10pt}
\paragraph{Watermarking for Diffusion Models}
Watermarking methods for diffusion models can be categorized into two paradigms: post-generation and in-generation. Post-generation methods embed watermark signals into images or videos after the content has been synthesized. Early approaches embed watermarks in the frequency domain~\citep{o1997rotation, cox1996secure, barni1998dct, kundur1998digital} or leverage SVD-based matrix operations~\citep{chang2005svd}. Recent post-generation methods typically employ deep neural networks to embed and extract watermark information from generated content~\citep{zhu2018hidden, zhang2019robust, zhang2023novel, fernandez2024video}.

On the other hand, in-generation methods embed the watermark into the images or videos during the generation process. \citet{wen2023tree} proposes Tree-ring that embeds a specific pattern into the initial Gaussian noise to achieve a 0-bit watermark, while Gaussian-shading~\citep{yang2024gaussian} embeds a multi-bit watermark into the initial Gaussian noise. Videoshield~\citep{hu2025videoshield} extends Gaussian shading to the video domain, marking the first in-generation watermarking method for video synthesis. However, these approaches require DDIM inversion to recover the initial Gaussian noise for watermark detection, resulting in extremely high computational cost for watermark insertion and extraction. Another line of work embeds watermarks by fine-tuning the latent decoder and employs an auxiliary decoder for watermark extraction, significantly reducing the overall runtime. For instance, \citet{fernandez2023stable} proposes Stable Signature, which finetunes the latent decoder of the LDMs to embed a multi-bit watermark message into the images. Similar to that, Lawa~\citep{rezaei2024lawa} and Wouaf~\citep{kim2024wouaf} fine-tune the latent decoder with an additional message encoder that embeds the multi-bit message into the images during generation. %\citet{zhang2024editguard} propose Editguard, which combines tamper localization with the watermark for diffusion models.
However, these image-based methods, when naively applied to video generation in a frame-wise manner, fail to account for the temporal consistency that is crucial to video data. In contrast to prior video watermarking approaches and naive frame-wise adaptations of image watermarking methods, our method bypasses the need for DDIMs inversion and embeds watermarks during generation by selectively fine-tuning perceptually insensitive layers of the latent decoder, while a temporal alignment module enforces consistency across frames.

\section{Methodology}

% \label{section: methodology}
In this section, we give a detailed description of Video Signature. Specifically, we give an overview of our training pipeline, and discuss how PAS works, and search the perceptually sensitive layers. Finally, we detail the training objective for each module.

\subsection{Overview of the Training Pipeline}
\label{section: overview}
Our training pipeline is shown in Figure~\ref{fig:pipeline}. We fine-tune the latent decoder \( \mathcal{D} \) to embed a multi-bit watermark message into each frame of the generated video. The latent encoder \( \mathcal{E} \), which is frozen during training, maps the input video \( \mathbf{v} \in \mathbb{R}^{f \times c \times H \times W} \) into a latent representation \( \mathbf{z} \in \mathbb{R}^{f \times c' \times h \times w} \), where \( f \) denotes the number of frames and \( c'\) denotes the channels of the latent space. The latent decoder \( \mathcal{D} \) then reconstructs the video as \( \hat{\mathbf{v}} = \mathcal{D}(\mathbf{z}) \in \mathbb{R}^{f \times c \times H \times W} \).
The reconstructed video \( \hat{\mathbf{v}} \) is subsequently fed to a pre-trained watermark extractor \( \mathcal{W} \), which extracts a fixed-length binary message from each frame, denoted as \( \hat{\mathbf{m}} = \mathcal{W}(\hat{\mathbf{v}}) \in \mathbb{R}^{f \times k} \), where \( k \) is the length of the embedded watermark per frame.  The ultimate goal of this training pipeline is to obtain a watermarked latent decoder \( \mathcal{D}' \), which can generate high-quality videos with imperceptible and robust watermarks directly, without additional modification during inference.
The goal of our training pipeline is summarized as:
\begin{equation}
\begin{split}
    \theta' = \theta + \delta, \quad
\delta^{\star} &= \arg\max_{\delta}\; \log p\!\bigl(\mathbf{m} \,\big|\, \mathcal{W}(\theta'(\mathbf{z}))\bigr)
\\
\text{s.t.}&\quad
\mathcal{P}\!\bigl(\theta(\mathbf{z}),\theta'(\mathbf{z})\bigr) \le \varepsilon,
\end{split}
\end{equation}
where \(\theta'\) and \(\theta\) denote the parameters of the watermarked and original decoder \(\mathcal{D}'\), \(\mathcal{D}\). Here \(\mathbf{z}\) is a latent vector, \(\textbf{m}\) is the watermark message, and \(\mathcal{P}\) is a perceptual distance metric (e.g., MSE) that evaluates the visual similarity between the watermarked video and its clean counterpart.

\subsection{PAS: Perturbation-Aware Suppression}
\label{scetion: PAS}

Existing watermarking work typically fine‑tunes \emph{all} decoder
parameters~\citep{fernandez2023stable, rezaei2024lawa, kim2024wouaf}, achieving high extraction
accuracy at the cost of perceptual artifacts. To mitigate this issue, we only optimize the layers with negligible impact to perceptual quality. We propose \textbf{Perturbation-Aware Suppression (PAS)}, which performs a straightforward but efficient search to identify layers with minimal perceptual impact on the output, enabling effective watermark embedding with minimal visual degradation. Specifically, for each layer \(L_j\) with parameters \(\theta_j\), we inject an isotropic Gaussian noise \(\epsilon \sim \mathcal{N}(0,\sigma^{2}I)\):
\begin{equation}
\theta_j' \;=\; \theta_j + \epsilon.
\end{equation}
Let \(\hat {\textbf{v}}_i^{(0)} = \mathcal{D}_{\theta}(\textbf{z}_i)\) be the reference output of
latent \(\textbf{z}_i\) and \(\hat{\mathbf{v}}_i^{(j)} = \mathcal{D}_{\theta'}(\textbf{z}_i)\) denotes the output after perturbing \(L_j\). The perceptual impact of layer \(L_j\) is then
estimated by:
\begin{equation}
s_j \;=\; \frac{1}{B}\sum_{i=1}^{B}
          \mathcal{P}\!\bigl(\hat {\mathbf{v}}_i^{(j)},\,\hat {\mathbf{v}}_i^{(0)}\bigr),
\label{eq:palf_score}
\end{equation}
where \(B\) is the number of latent samples.  Finally, the layer set for
fine‑tuning is selected as \(\mathcal{L}_{\mathrm{ft}} \;=\; \{\,L_j \mid s_j < \tau\,\}\), with a threshold \(\tau\).  We detail this in Algorithm~\ref{alg:pas}.

\begin{algorithm}[tb]
\caption{Perturbation-Aware Suppression}
\label{alg:pas}
\textbf{Input}: Latent batch \(\{\mathbf{z}_i\}_{i=1}^B\)\\
\textbf{Parameter}: Frozen Decoder \( D = \{L_1, \ldots, L_n\} \),  Layer Parameter \( \theta = \{\theta_1, \ldots, \theta_n\}\), Perceptual distance metric \( \mathcal{P}(\cdot, \cdot) \), Noise scale \(\sigma\), Threshold \(\tau\)\\
\textbf{Output}: Selected layer set \( \mathcal{L}_{\text{ft}} \)
\begin{algorithmic}[1] %[1] enables line numbers
\STATE Generate reference outputs \( \hat{\mathbf{v}}_i^{(0)} = \mathcal{D}(\mathbf{z}_i) \) for all \( \mathbf{z}_i \).

\FOR{each layer \( L_j \in \mathcal{D} \)}
   \STATE \( \theta_j^{ori} \leftarrow \theta_j\)
   \STATE \( \theta_j \leftarrow \theta_j + \epsilon, \quad \epsilon \sim \mathcal{N}(0, \sigma^2) \)
    \FOR{each \( \mathbf{z}_i \) in batch}
        \STATE \( \hat{\mathbf{v}}_i^{(j)} \gets \mathcal{D}(\mathbf{z}_i) \)
        \STATE \( d_i^{(j)} \gets \mathcal{P}(\hat{\mathbf{v}}_i^{(j)}, \hat{\mathbf{v}}_i^{(0)}) \)
    \ENDFOR
    \STATE \( s_j \gets \frac{1}{B} \sum_{i=1}^B d_i^{(j)} \)
    \STATE \( \theta_j \leftarrow \theta_j^{ori}\)
\ENDFOR
\STATE \( \mathcal{L}_{\text{ft}} \gets \{L_j \mid s_j < \tau \} \)
\STATE \textbf{Return} \( \mathcal{L}_{\text{ft}} \)
\end{algorithmic}
\end{algorithm}

\subsection{Training Objectives}

\label{section:training objective}
As illustrated in Figure~\ref{fig:pipeline}, our training framework is designed to embed a binary watermark into the video generation process while preserving high perceptual quality and temporal consistency. The overall optimization involves two objectives: the watermark extraction and the visual alignment in the spatial and temporal domains.
\vspace{-10pt}
\paragraph{Watermark Extraction} 
To ensure accurate watermark embedding across the entire video, we employ two complementary loss terms: a frame-wise watermark loss and a video-level watermark loss. Given a target binary watermark message \( \mathbf{m} \in \{0,1\}^k \), the watermarked video \( \hat{\mathbf{v}} \in \mathbb{R}^{f \times c \times H \times W} \) is generated by decoding latent vector \( \mathbf{z} \) through the fine-tuned decoder \( \mathcal{D}' \). A pretrained extractor \( \mathcal{W} \) predicts a soft watermark vector \( \hat{\mathbf{m}}_t \in [0,1]^k \) from each frame \( \hat{\mathbf{v}}_t \).
We directly supervise each predicted frame-wise message with the target watermark via binary cross entropy:
\begin{equation}
\mathcal{L}_{\text{fr}} = -\frac{1}{f} \sum_{t=1}^{f} \sum_{i=1}^{k} \left[ m_i \log \hat{m}_{t,i} + (1 - m_i) \log (1 - \hat{m}_{t,i}) \right].
\end{equation}
To improve the global consistency of watermark extraction, we aggregate predictions across all frames to form a soft message vector and compute a global-watermark loss:
\begin{equation}
\mathcal{L}_{\text{v}} = - \sum_{i=1}^{k} \left[ m_i \log \bar{m}_{i} + (1 - m_i) \log (1 - \bar{m}_{i}) \right],
\end{equation}
\begin{table*}[t]
\small
\centering
\setlength{\tabcolsep}{1.95mm}
\renewcommand{\arraystretch}{0.95}
\caption{Performance comparison of watermarking methods on Modelscope (MS) and SVD. Video Quality refers to the average value of the four metrics from VBench. PSNR, SSIM, LPIPS, and tLP are calculated between the watermarked video and the vanilla video (non-watermarked video), VideoShield is not applicable for these metrics since there are no vanilla videos for this kind of methods. \textbf{Bold} indicates the best result within each model and method type (post-generation or in-generation). Arrows denote whether higher (\(\uparrow\)) or lower (\(\downarrow\)) values indicate better performance. Theoretically, the \(T_i\) and \(T_e\) of the same method should remain consistent across models. However, since the experiments were run independently, small variations are observed.}
\begin{tabular}{c|lcccccccc}
\toprule
Model & Method & Bit Accuracy \(\uparrow\) & \(T_i \downarrow\) & \(T_e \downarrow\) & PSNR \(\uparrow\) & SSIM \(\uparrow\) & LPIPS \(\downarrow\) & tLP \(\downarrow\) & Video Quality \(\uparrow\) \\
\midrule

\multirow{7}{*}{MS}
  & \multicolumn{9}{l}{\textit{Post-generation methods}} \\
  & RivaGAN                & 0.938         & 0.711  & 0.159  & 34.973 & 0.947 & 0.117 & 0.012    & 0.893 \\
  & VideoSeal              & \textbf{0.975}              & \textbf{0.121}  & \textbf{0.036}  & \textbf{35.113}    & \textbf{0.947} & \textbf{0.083}  & \textbf{0.009} & \textbf{0.893} \\
  \cmidrule(l){2-10}
  & \multicolumn{9}{l}{\textit{In-generation methods}} \\
  & StableSig              & 0.995  & 0.000  & 0.008  & 29.448             & 0.799              & 0.165 & \textbf{0.008}  & 0.893 \\
  & VideoShield           & \textbf{1.000}     & 26.994 & 6.224  & -             & -              & - & -  & \textbf{0.894} \\
  & \textsc{VidSig} (Ours)  & 0.992              & \textbf{0.000}  & \textbf{0.008}  & \textbf{30.523}             & \textbf{0.840}              & \textbf{0.151} & 0.009 & 0.893 \\

\midrule

\multirow{7}{*}{SVD}
  & \multicolumn{9}{l}{\textit{Post-generation methods}} \\
  & RivaGAN          & 0.886              & 0.686  & 0.162  & 35.561 & 0.963 & 0.069 & 0.004 & 0.871 \\
  & VideoSeal        & \textbf{0.979}     & \textbf{0.049}  & \textbf{0.018}  & \textbf{35.887}    & \textbf{0.966}   & \textbf{0.060} & \textbf{0.003} & \textbf{0.871} \\
  \cmidrule(l){2-10}
  & \multicolumn{9}{l}{\textit{In-generation methods}} \\
  & StableSig        & 0.998  & 0.000  & 0.005  & 30.080             & 0.906              & 0.100 & 0.003 & 0.873 \\
  & VideoShield      & 0.990              & 28.586 & 19.813 & -             & -              & - & - & 0.867 \\
  & \textsc{VidSig} (Ours)  & \textbf{0.999} & \textbf{0.000}  & \textbf{0.005}  & \textbf{31.662}     & \textbf{0.924} & \textbf{0.091} & \textbf{0.003} & \textbf{0.873} \\
\bottomrule
\end{tabular}

\label{tab:main results}
\end{table*}
where \(\bar{\mathbf{m}} = \frac{1}{f} \sum_{t=1}^{f} \hat{\mathbf{m}}_t\). Eventually, the final watermark loss is a weighted sum \(\mathcal{L}_{\text{wm}} = \alpha_1 \mathcal{L}_{\text{fr}} + \alpha_2 \mathcal{L}_{\text{v}}\),    
where \( \alpha_1, \alpha_2 > 0 \) are hyperparameters balancing local (frame-level) and global (video-level) watermarks to enhance the watermark extraction performance. In this paper, we set \( \alpha_1 =  \alpha_2 =1 \) as default. During inference, the extractor first predicts the bit message for each frame, then obtains the watermark for the entire video by majority voting.
\vspace{-10pt}
\paragraph{Visual Alignment} To ensure the watermarked video remains visually similar to the non-watermarked video, we conduct a spatial alignment between the frame-wise outputs of the original decoder \( \mathcal{D} \) and the watermarked decoder \( \mathcal{D}' \). Let \( \mathbf{v} \) and \( \hat{\mathbf{v}} \) be the reconstructed non-watermarked and watermarked videos, respectively, both of shape \( f \times c \times H \times W \). The perceptual loss in the spatial domain is defined as \(
\mathcal{L}_{\text{spatial}} = \frac{1}{f} \sum_{t=1}^f \mathcal{D}_{\text{sim}}(\hat{\mathbf{v}}_t, \mathbf{v}_t),     
\)
where \( \mathcal{D}_{\text{sim}}(\cdot, \cdot) \) denotes a generic frame-level similarity metric, such as MAE, MSE, and LPIPS~\citep{zhang2018unreasonable}. In this paper, we use Watson-VGG~\citep{czolbe2020loss}, an improved version of LPIPS, as the perceptual loss.

To further align the visual similarity in the temporal domain, we introduce a straightforward but effective module, which we refer to as \textbf{Temporal Alignment}. We simply align the \textbf{Inter-Frame Dynamics} between \( \mathbf{v} \) and \( \hat{\mathbf{v}} \). Specifically, we compute frame-wise differences \( \Delta_t = \mathbf{v}_{t+1} - \mathbf{v}_t \) and \( \hat{\Delta}_t = \hat{\mathbf{v}}_{t+1} - \hat{\mathbf{v}}_t \), as shownd in Figure~\ref{fig:pipeline} and define:
\begin{equation}
\mathcal{L}_{\text{temporal}} = \frac{1}{f-1} \sum_{t=1}^{f-1} \mathcal{D}_{\text{sim}}(\hat{\Delta}_t, \Delta_t).
\end{equation}
We continue to use Watson-VGG as the distance metric for temporal alignment. Additional evaluation results based on other metrics are reported in Section~\ref{ablation study}. Thus, the total training loss is a weighted combination of the three objectives:
\begin{equation}
\mathcal{L} = \lambda_1 \mathcal{L}_{\text{wm}} + \lambda_2 \mathcal{L}_{\text{spatial}} + \lambda_3 \mathcal{L}_{\text{temporal}},    
\end{equation}
where \( \lambda_1 \), \( \lambda_2 \), and \( \lambda_3 \) are hyperparameters controlling the trade-off between the watermark accuracy and the fidelity of the video. We set \( \lambda_1 = 1 \), and \( \lambda_2 = \lambda_3 = 0.2 \) as default.

% \section{Final copy}

% You must include your signed IEEE copyright release form when you submit your finished paper.
% We MUST have this form before your paper can be published in the proceedings.

% Please direct any questions to the production editor in charge of these proceedings at the IEEE Computer Society Press:
% \url{https://www.computer.org/about/contact}.
\section{Experiment}

\label{section: EXP}
\subsection{Experiment Settings}
\label{section: Exp settings}
\paragraph{Implementation Details}
We evaluate our method on two video generation models: the text-to-video (T2V) model ModelScope (MS)~\citep{wang2023modelscope}, and the image-to-video (I2V) model Stable Video Diffusion (SVD)~\citep{blattmann2023stable}. Before training, we sample 10 vanilla generated videos to run our proposed PAS for 10 iterations to get the perceptual-sensitive layers and freeze them. During training, we fine-tune only the remained layers of the latent decoder after PAS. In addition. We use the pretrained watermark extractor provided by~\citet{fernandez2023stable}, which can extract a 48-bit binary message from a single image. Optimization is performed using the AdamW optimizer~\citep{loshchilov2017decoupled}, with a learning rate of \( 5 \times 10^{-4} \). We run all of our experiments on one NVIDIA A800 GPU. We give additional implementation details in the Appendix.

\vspace{-10pt}
\paragraph{Datasets}
We use the OpenVid-1M dataset~\citep{nan2024openvid} for training. Specifically, we randomly sample 10{,}000 videos from that to fine-tune the latent decoder. Each training video contains 8 frames sampled at a frame interval of 8. For evaluation, we select 50 prompts from the test set of VBench~\citep{huang2024vbench}, covering five categories: Animal, Human, Plant, Scenery, and Vehicles (10 prompts each). For the T2V task, we generate four videos per prompt using four fixed seeds, resulting in 200 videos. For the I2V task, we first generate images using Stable Diffusion 2.1~\citep{rombach2022high} based on the same prompts, and generate videos conditioned on these images using the same fixed seeds. In total, for both T2V and I2V tasks, we generate 200 videos, each consisting of 16 frames with a resolution of \(512 \times 512\). The inference steps of the two models are set to 25 and 50, respectively.
\begin{figure}[t]
    \centering
    \includegraphics[width=0.95\linewidth]{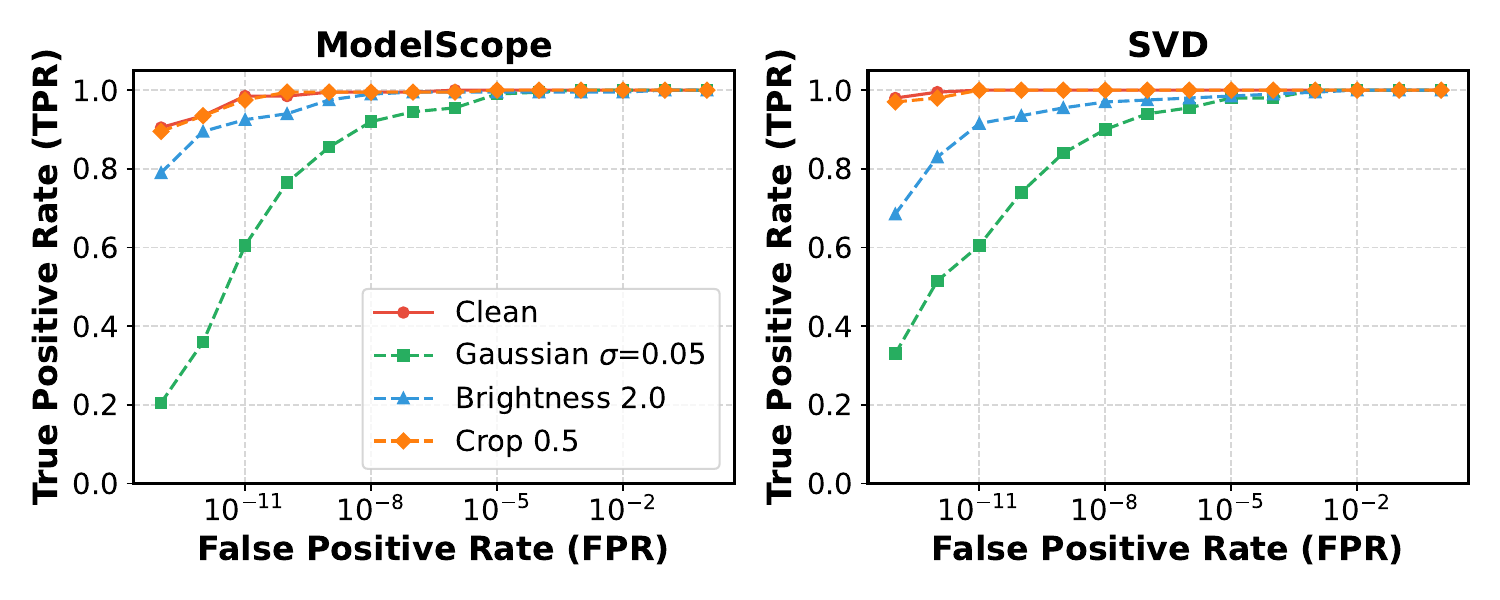}
    \caption{Watermark detection of Video Signature.}
    \vspace{-15pt}
    \label{fig:tpr}
\end{figure}

\vspace{-10pt}
\paragraph{Baseline} We compare our method with 4 watermarking methods: RivanGAN~\citep{zhang2019robust}, VideoSeal~\citep{fernandez2024video}, Stable Signature~\citep{fernandez2023stable}, VideoShiled~\citep{hu2025videoshield}. Among these methods, RivanGAN and VideoSeal are post-generation methods, Stable Signature is a train-based in-generation method for image watermarking, and VideoShield is a train-free in-generation method for video watermarking.
\begin{table*}[t]
\centering
\small
\caption{Bit Accuracy under different kinds of temporal tampering: FD (Frame Drop), FS (Frame Swap), FI (Frame Insert), FIG (Frame Insert Gaussian), FA (Frame Average, where n refers to the average length). The final column denotes the average performance. We give detailed information of implementations of these tampering methods in our Appendix. Note that the detection of VideoShield only works when the number of frames in the generated video is exactly equal to that in the tampered video. \textbf{Bold} indicates the best performance, and \underline{underlined} indicates the second best.}
\begin{tabular}{c|lcccccccccccc}
\toprule
Model & Method & Benign & FD & FS &FI & FIG & \makecell{FA (n=3)} & \makecell{FA (n=5)} & \makecell{FA (n=7)} & \makecell{FA (n=9)} & Avg\\
\midrule
\multirow{5}{*}{MS}
  & RivaGAN          &   0.938   & 0.938 & 0.938 & 0.938 &0.935 & 0.938 & 0.938 & 0.937 & 0.936 &  0.937 \\
  & VideoSeal         & 0.975	 & 0.974  & 0.975 & 0.974  & \underline{0.975} & 0.972 & 0.972 & 0.967 & 0.958  &   0.971 \\
  & StableSig        &\underline{0.995}& \textbf{0.995} & \underline{0.995} & \textbf{0.995} & 0.964& \textbf{0.992} & \underline{0.988} & \underline{0.984} & \underline{0.980} &\underline{0.988} \\   
  & VideoShield      & \textbf{1.000}   & - & \textbf{1.000} & - &  - & - & - & - &  - & - \\
  & \textsc{VidSig} (ours) &  0.992 &\underline{0.992} & 0.992 & \underline{0.992} & \textbf{0.992} & \textbf{0.992} & \textbf{0.992} & \textbf{0.992} & \textbf{0.992} & \textbf{0.992}\\
\midrule
\multirow{5}{*}{SVD} 	 	
  & RivaGAN          & 0.886	      & 0.886 & 0.886  &  0.885 & 0.881 & 0.883 & 0.882 & 0.880 & 0.873 & 0.882   \\
  & VideoSeal        & 0.979	      & 0.977  & 0.979 & 0.978 &\underline{0.979} & 0.975 & 0.973 & 0.971 & 0.965    & 0.975    \\
  & StableSig        & \underline{0.998}  & \underline{0.998}  & \underline{0.998}  & \underline{0.998} &0.966 & \underline{0.995} & \underline{0.989} & \underline{0.980} & \underline{0.972} & \underline{0.988} \\
  & VideoShield      & 0.990 & - & 0.964 & -  & - & - & - & - & - & - &     \\
  & \textsc{VidSig} (ours) & \textbf{0.999} & \textbf{0.999} & \textbf{0.999} & \textbf{0.999} & \textbf{0.999} & \textbf{0.999} & \textbf{0.999} & \textbf{0.999} & \textbf{0.999} &\textbf{0.999}\\ 
\bottomrule
\end{tabular}
\label{tab:robustness}
\end{table*}

\vspace{-10pt}
\paragraph{Metrics} We use the True Positive Rate (TPR) corresponding to a fixed False Positive Rate (FPR) to assess the watermark detection. Meanwhile, we use Bit Accuracy to evaluate the watermark extraction accuracy. To assess the visual quality of generated videos, we leverage four official evaluation metrics from VBench~\citep{huang2024vbench}, including: Subject Consistency, Background Consistency, Motion Smoothness, and Imaging Quality. We also use four standard perceptual quality metrics for evaluating the distortion of the generated videos: PSNR, SSIM~\citep{wang2004image}, LPIPS~\citep{zhang2018unreasonable}, and its extension tLP~\citep{chu2020learning}. We provide the details of them in the Appendix. To assess the efficiency of watermarking methods, we evaluate the insertion and extraction times, denoted as \(T_i\) and \(T_e\). The insertion time \(T_i\) is defined as the additional time cost introduced by watermarking, measured as the difference in runtime between vanilla generation and watermarked video generation. The extraction time \(T_e\) refers to the time required to recover the watermark from a generated video.

\subsection{Main Results}
\label{section: main results}
\paragraph{Watermark Detection}

% We fix the false positive rate (FPR) from \(10^{-13}\) to \(10^0\) and compute the true positive rate (TPR). the results are shown in Figure~\ref{fig:tpr}. It's seen that \textsc{VidSig} maintains an almost perfect TPR across all tampering types even at extremely low FPR (\(10^{-11}\)), and achieves a TPR above 0.95 under additive Gaussian noise with FPR = \(10^{-6}\), it also achieves a TPR at 0.995 under FPR = \(10^{-4}\). 
Figure~\ref{fig:tpr} illustrates the detection performance of \textsc{VidSig}. The method maintains a near-perfect True Positive Rate (TPR) across most tampering types, even at an extremely low False Positive Rate (FPR) of $10^{-11}$. Its robustness is further demonstrated under the challenging additive Gaussian noise attack; \textsc{VidSig} achieves a high TPR of 0.995 at FPR = \(10^{-4}\), and still maintains a TPR above 0.95 at the even stricter FPR = \(10^{-6}\).

\vspace{-10pt}
\paragraph{Comparison to Baselines}
Table~\ref{tab:main results} presents the performance of \textsc{VidSig} and other baseline methods. The findings are summarized as follows:
\textbf{(i)} Post-generation methods deliver the poorest extraction accuracy on both tasks, even their best case, VideoSeal on SVD, reaches only 0.979 and confers no perceptual benefit, underscoring an unfavorable accuracy–fidelity trade‑off.
\textbf{(ii)} By contrast, in-generation methods achieve a nearly perfect (up to 1 for T2V and 0.999 for I2V) extraction accuracy in both tasks and maintain a reasonable visual quality. The PSNR, SSIM, etc, are lower, but the VBench scores are comparable and even higher.
\textbf{(iii)} In-generation methods, except for VideoShield, achieve a negligible latency for watermark embedding and the lowest cost for watermark detection.

\textsc{VidSig} surpasses other baseline methods by \textbf{(i)} matching or exceeding their near‑perfect bit accuracy (0.992 for T2V  and 0.999 for I2V) and achieves the best or near‑best VBench scores on both tasks. \textbf{(ii)} The highest PSNR/SSIM and the lowest LPIPS/tLP values among in-generation methods reveal that \textsc{VidSig} introduces minimal visual artifacts compared to other in-generation methods. \textbf{(iii)} \textsc{VidSig} attains the lowest insertion and extraction computational overhead of all methods and, while matching Stable Signature in runtime, surpasses it on nearly every evaluation metric.

\begin{table}[t]
\centering
\small
\renewcommand{\arraystretch}{1}
\setlength{\tabcolsep}{2.5mm}
\caption{Watermark detection of the videos generated by Stable Video Diffusion, under H.264 compression in real-world scenario.}
\resizebox{\linewidth}{!}{
\begin{tabular}{c|cccc}
\toprule
H.264  & BA & TPR@1e-4 & TPR@1e-3 & TPR@1e-2 \\
\midrule
CRF = 20 & 0.894 & 0.880 & 0.895 & 0.945 \\
CRF = 23 & 0.869 & 0.840 & 0.855 & 0.880 \\
\bottomrule
\end{tabular}
}
\label{tab:H264}
\end{table}

%Configuration & Bit Accuracy & PSNR  & SSIM  & LPIPS & tLP  & Video Quality  \\

 \begin{figure*}[t]
    \centering
    \includegraphics[width=\textwidth]{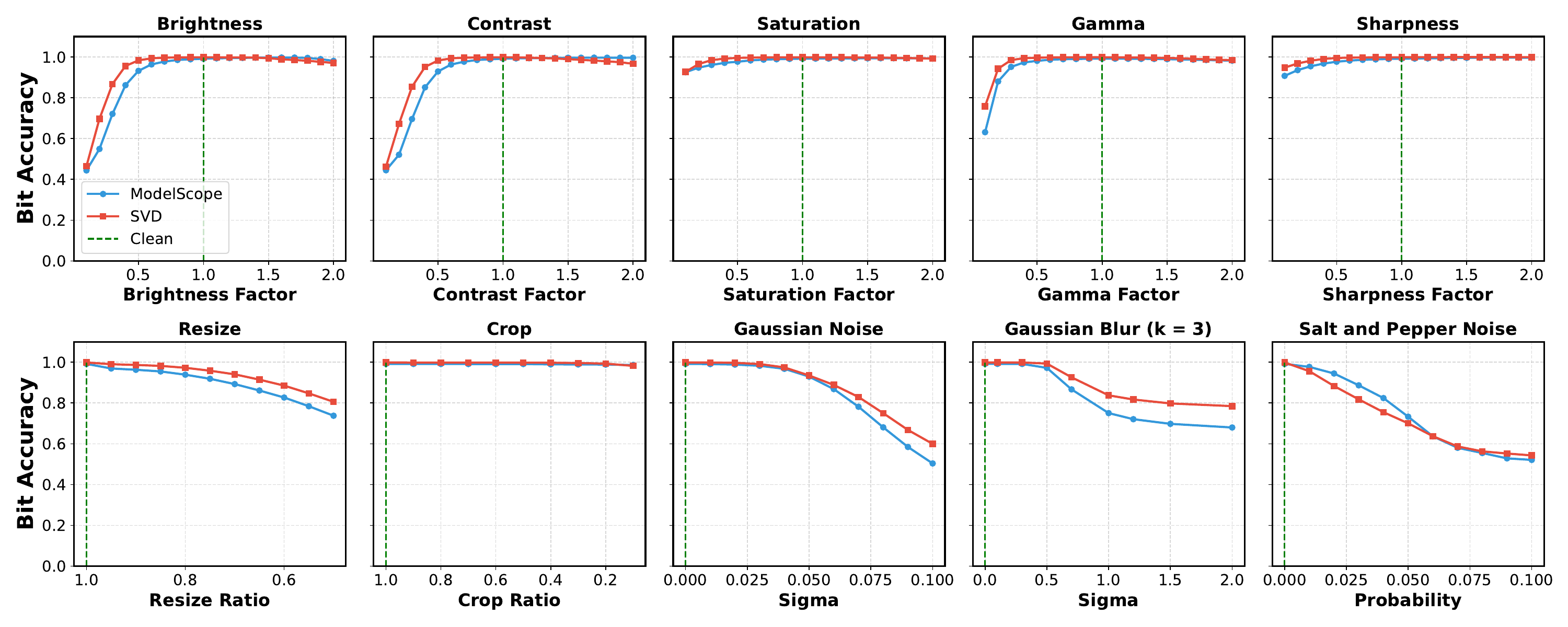}
    \caption{Bit accuracy under different spatial tampering. The attack is applied to each frame.}
    \label{fig:robustness}
\end{figure*}

% \begin{table}[t]
% \centering
% \small
% \setlength{\tabcolsep}{1.9mm}
% \renewcommand{\arraystretch}{1}
% \caption{Watermark Detection under H.264 compression in real-world scenario.}
% \begin{tabular}{c|cc}
% \toprule
% H.264& Bit Accuracy  & TPR  \\

% \midrule
% CRF = 20 & 0.894 & 0.9  \\
% CRF = 23 & 0.867 & 0.910  \\
% \bottomrule
% \end{tabular}
% \label{tab:transfer}
% \end{table}

\subsection{Robustness}
\label{section:robustness}

\paragraph{Temporal Tamper}Table~\ref{tab:robustness} presents the extraction accuracy of different watermarking methods under several temporal perturbations and the benign setting. Overall, \textsc{VidSig} consistently achieves near-perfect extraction accuracy across all attacks. For T2V task, \textsc{VidSig} matches or closely follows the best-performing method (VideoShield) while requiring significantly less computational cost (see Table~\ref{tab:main results}). It's noticed that the watermark detection of VideoShield strictly requires the video to remain the same resolution and number of frames after temporal tamper, which limits its application.
For the I2V task, \textsc{VidSig} achieves the highest extraction accuracy (0.999) across all perturbations, outperforming all baselines. This resilience to temporal tampering chiefly arises from embedding watermarks in every frame and aggregating the detections through majority voting.

Besides the above attacks, we further investigate whether \textsc{VidSig} remains robust in real-world media. We compress the videos generated by Stable Video Diffusion using the H.264 codec and evaluate detection under different FPR levels; the results are reported in Table~\ref{tab:H264}. It can be observed that video compression leads to a moderate drop in bit accuracy and TPR, especially under stronger compression (larger CRF). Nevertheless, both BA and TPR stay at a high level across all settings, indicating that \textsc{VidSig} still provides reliable watermark detection after practical H.264 compression and is suitable for real-world deployment.

\vspace{-10pt}
\paragraph{Spatial Tamper}We further evaluate the resilience of \textsc{VidSig} against various types and intensities of spatial tampering, as illustrated in Figure~\ref{fig:robustness}. We find that \textsc{VidSig} remains effective under most perturbations. In particular, it maintains robust performance within a certain range when subjected to Gaussian noise, Gaussian blur, and Salt and Pepper Noise. It is worth noting that no additional robustness enhancement techniques were applied during fine-tuning. These results demonstrate the robustness of \textsc{VidSig} against both temporal and spatial tampering, highlighting its potential for real-world deployment.

\section{Analysis}
\label{sec: analysis}
\subsection{Impact of Generation Configuration}

\begin{figure}[tb]
  \centering
  \includegraphics[width=0.95\linewidth]{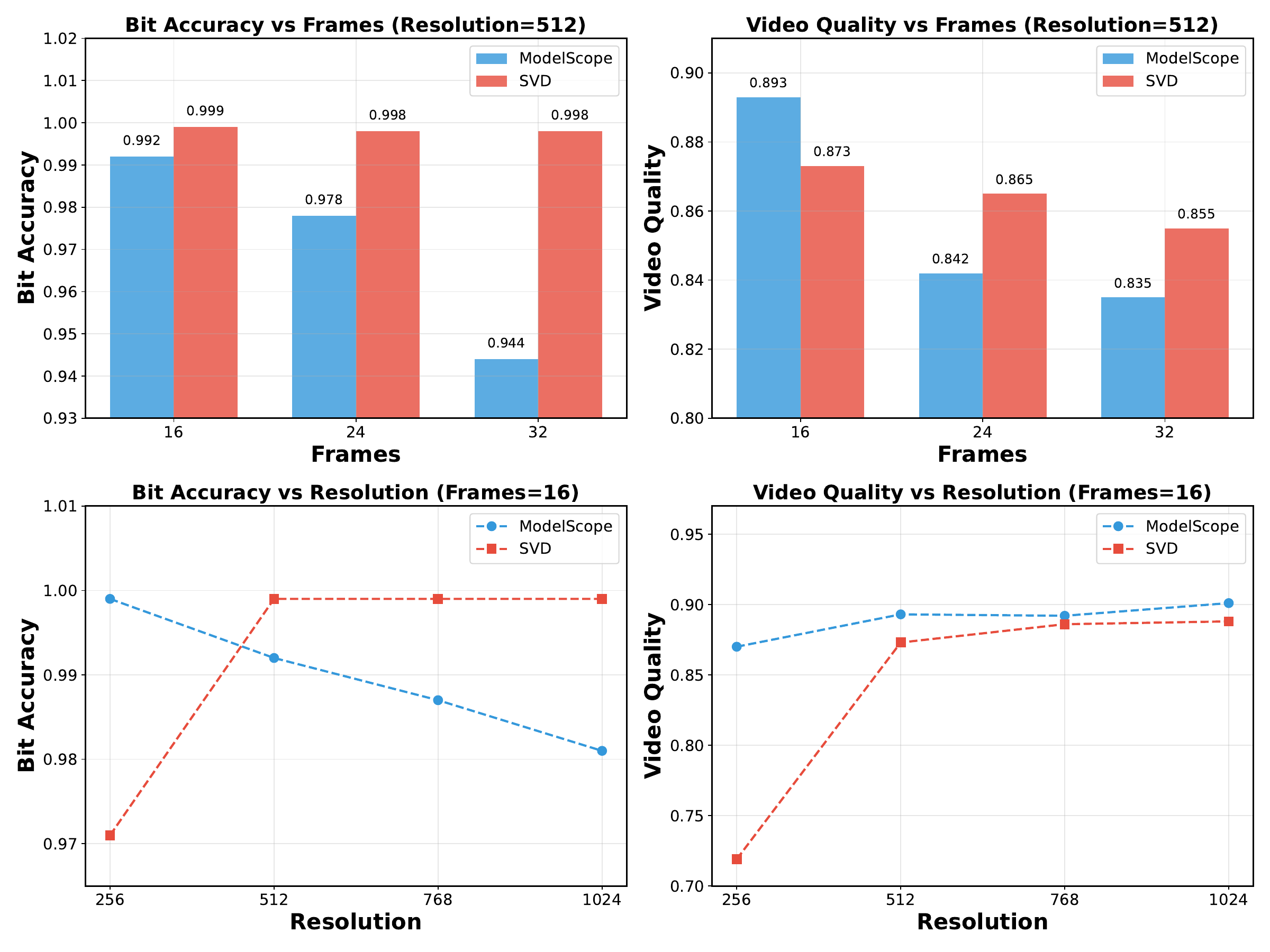}
  \caption{Extraction Accuracy and Video Quality versus Frames and Resolution. Resolution \(= N\) denotes to a resolution of \(N \times N\).}
  \label{fig:config}
\end{figure}

\begin{table}[t]
\centering
\caption{Performance of \textsc{VidSig} in Latte transferred from ModelScope (MS) and SVD. The inference step is set to 50, prompts and generators follow the same settings in~\ref{section: Exp settings}. The True Positive Rate is reported under \(\text{FPR} = 1\times10^{-6}\).}
\small
\setlength{\tabcolsep}{1.9mm}
\renewcommand{\arraystretch}{1}
\begin{tabular}{c|cccccc}
\toprule
Decoder & BA  & TPR  & PSNR  & SSIM  & LPIPS  & tLP \\

\midrule
\(\text{Latte}_{\textit{MS}}\)  & 0.998 & 1.000 & 31.073 & 0.875 & 0.133 & 0.007 \\
\(\text{Latte}_{\textit{SVD}}\) & 0.998 & 1.000 & 29.564 & 0.839 & 0.148 & 0.020 \\
\bottomrule
\end{tabular}
\label{tab:transfer}
\end{table}

We conduct experiments to investigate how generation configurations affect watermark extraction performance and the resulting video quality. Specifically, we vary the number of generated frames and the output resolutions while keeping other configurations, such as the inference step, unchanged. The results are shown in Figure~\ref{fig:config}.
\vspace{-10pt}

\paragraph{Impact of Frames}
When the number of generated frames increases, the watermark extraction accuracy decreases slightly for the T2V model (ModelScope), from 0.992 to 0.944, but it still remains a high TPR, about 0.965 under FPR \(=1e-4\). The video quality drops by about 0.06 when the frames increase from 16 to 32. In contrast, the I2V model (SVD) remains very stable as the number of frames grows, with only a minor decrease in video quality. We also observe that this quality drop is mainly caused by the generation model itself rather than the watermark in the additional frames, since the average difference between the non-watermarked and watermarked videos, denoted as \(\bar\Delta_{vq}\), is close to 0 in our experiments. The complete numerical comparison of video quality across the original and watermarked models is provided in the Appendix.

\vspace{-10pt}
\paragraph{Impact of Resolution}
When the resolution increases from \(256 \times 256\) to \(1024 \times 1024\), the extraction accuracy of the T2V model continues to decrease but stays above 0.980. In contrast, the extraction accuracy of the I2V model increases from 0.971 to 0.999 and maintains almost perfect accuracy. At the same time, the video quality of both the T2V and I2V models improves as the resolution becomes higher. These findings prove that \textsc{VidSig} can remain stable across different video lengths and resolutions.

\subsection{Transferability}

\label{sec: EXP transfer}

\begin{figure}[tb]
  \centering
  \includegraphics[width=0.95\linewidth]{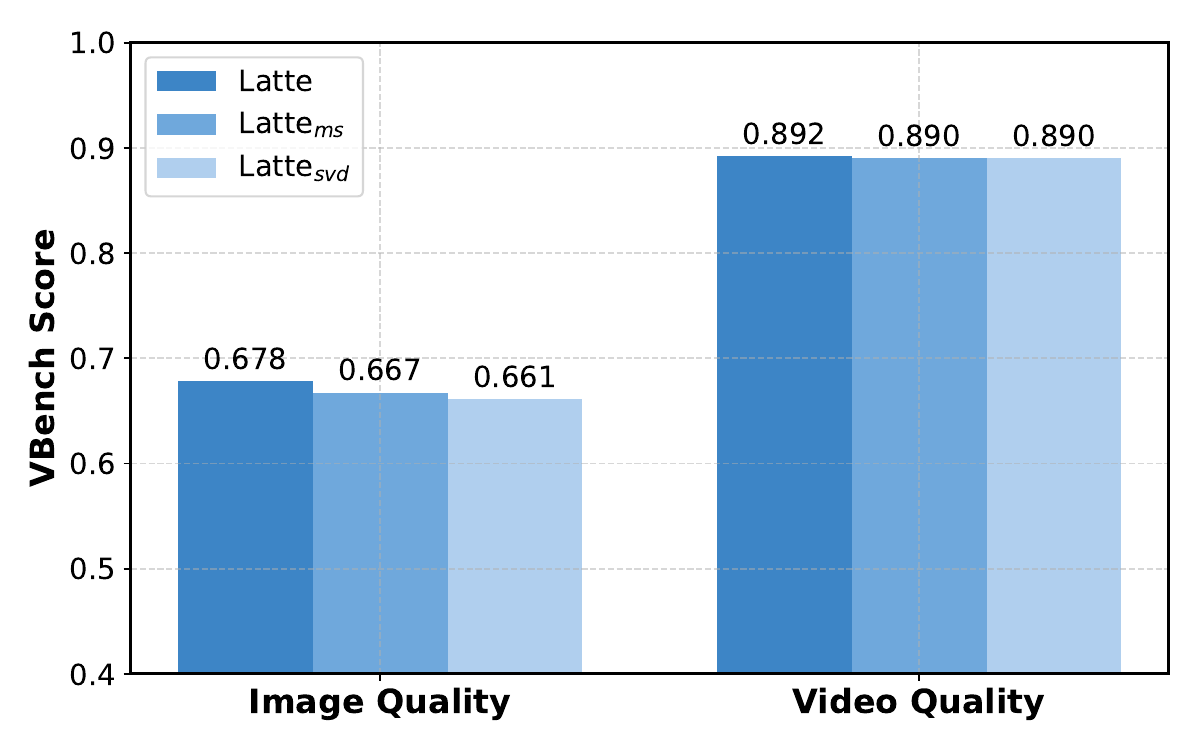}
  \caption{VBench score of \textsc{VidSig} in Latte transferred from MS and SVD. Latte refers to the original model. Image Quality is one of the four metrics we have mentioned in Section~\ref{section: Exp settings}, and Video Quality is the same as that in Table~\ref{tab:main results}.}
  \label{fig:transfer}
\end{figure}

We further investigate the transferability of \textsc{VidSig}. We transfer the fine-tuned latent decoder of ModelScope and SVD to a completely new model and evaluate whether the embedded watermark can still be successfully extracted. Specifically, we substitute the original latent decoder of Latte~\citep{ma2024latte}, a DiT-based model, with our fine-tuned watermarked latent decoder. The results in Table~\ref{tab:transfer} and Figure~\ref{fig:transfer} indicate that \textsc{VidSig} achieves high extraction accuracy and a perfect TPR even in cross-model scenarios, demonstrating its strong transferability. This property suggests that the fine-tuned latent decoder can serve as a \textbf{plug-in} module in most of the current video diffusion models, making \textsc{VidSig} more versatile in real-world deployment.

\begin{table}[t]
\centering
\small
\setlength{\tabcolsep}{1mm}
\caption{Ablation results of PAS and TA modules on ModelScope with consistent experiment settings. \textbf{Bold} and \underline{underlined} indicate the best and the second best performance.}
\begin{tabular}{c|cccccc}
\toprule
Configuration & BA  \(\uparrow\) & PSNR \(\uparrow\) & SSIM \(\uparrow\) & LPIPS \(\downarrow\) & tLP \(\downarrow\) & VQ \(\uparrow\) \\

\midrule
w/o ALL     & 0.990 & 28.565 & 0.786 & 0.182 & 0.011 & 0.889 \\
w/ TA       & 0.991  & \underline{29.793}  & \underline{0.815}  & 0.164  & 0.010  & \textbf{0.894}  \\
w/ PAS      & \textbf{0.993}  & 29.762 & 0.814  & \underline{0.163}  & \textbf{0.009}  & 0.892  \\
w/ ALL      & \underline{0.992}  & \textbf{30.523}  & \textbf{0.840}  & \textbf{0.151}  & \textbf{0.009}  & \underline{0.893}  \\
\bottomrule
\end{tabular}

\label{tab:module_ablation}
\end{table}

\subsection{Ablation Study}
\label{ablation study}

\paragraph{Effectiveness of PAS and TA} Table~\ref{tab:module_ablation} presents the ablation study evaluating the contributions of the two key components in our framework: Perturbation-Aware Suppression (PAS) and Temporal Alignment (TA). We can see that each component individually yields substantial gains across all evaluation metrics, and their integration attains the highest overall performance—with only a marginal decrease in bit accuracy—thereby confirming that PAS and TA are complementary and jointly indispensable for realizing the full potential of the proposed framework.

\begin{figure}[t]
     \centering
     \includegraphics[width=0.95\linewidth]{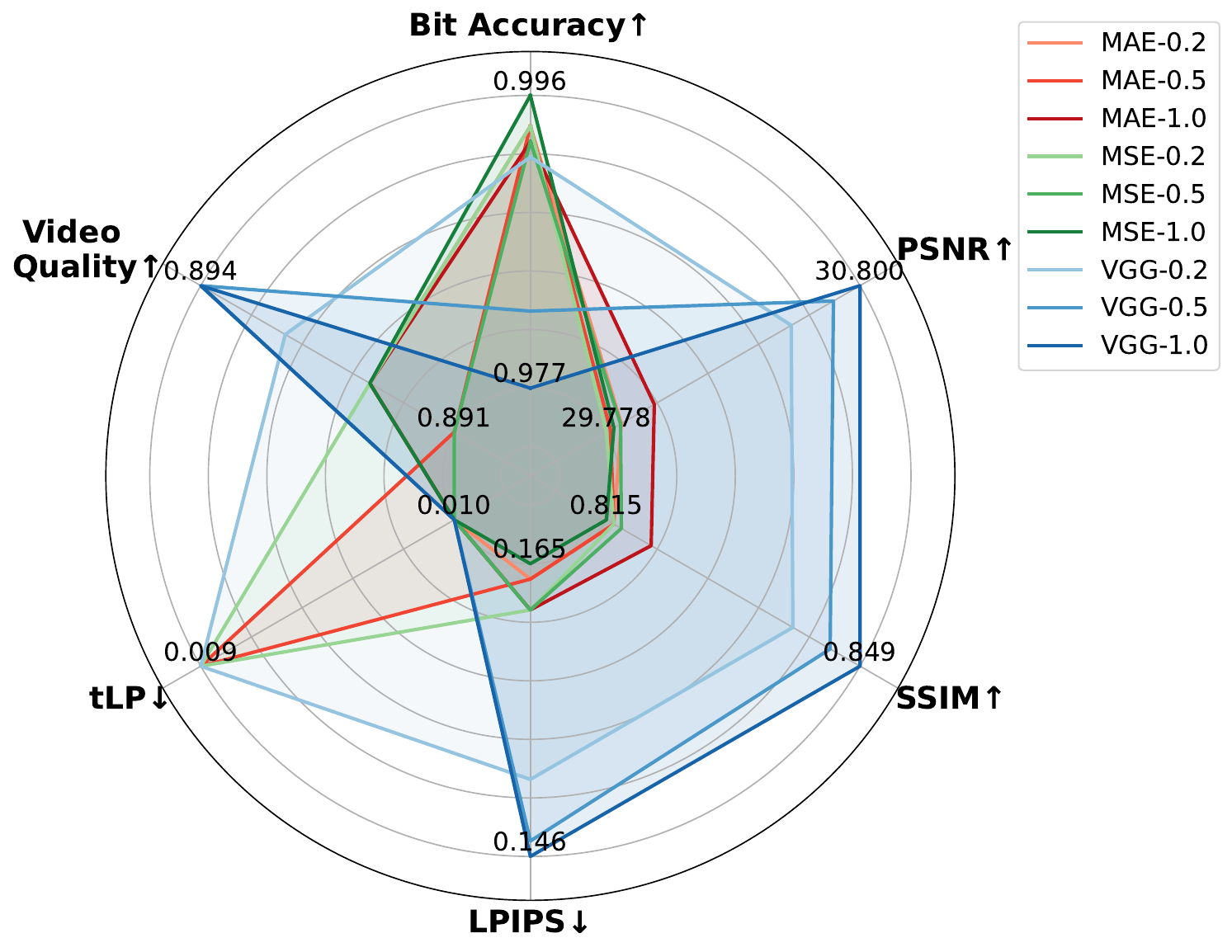}
     \caption{Ablation study on different temporal alignment metrics under varying $\lambda_3$ values. Color hues represent metric type, and color intensity indicates increasing $\lambda_3$.}
     \label{fig:ablation_radar}
 \end{figure}

\vspace{-10pt}
\paragraph{Different Perceptual Metric for TA} We then evaluate three different temporal alignment strategies, MAE, MSE, and Watson-VGG (we use VGG for abbreviation in the following discussion), to identify the most effective formulation for preserving temporal consistency. The results are shown in Figure~\ref{fig:ablation_radar}. It's seen that MAE and MSE produce slightly higher bit accuracy yet noticeably degrade video quality. In contrast, the VGG loss with \(\lambda_{3}=0.2\) attains comparable bit accuracy while substantially improving perceptual quality, thereby offering the most favorable overall trade-off across all metrics. We attribute this superiority to the smoother, perceptually informed gradients supplied by the VGG loss during fine-tuning.

\section{Conclusion}

In this paper, we propose Video Signature (\textsc{VidSig}), an in-generation watermarking method for latent video diffusion models, which embeds robust and imperceptible watermarks during the generation process. Our framework introduces Perturbation-Aware Suppression (PAS) and Temporal Alignment (TA), enabling \textsc{VidSig} to achieve state-of-the-art watermark extraction accuracy while maintaining high visual fidelity and minimal computational overhead. Beyond outperforming existing methods, \textsc{VidSig} demonstrates strong robustness against both spatial and temporal tamper and even shows transferability across models. These results highlight its potential for real-world deployment in protecting the intellectual property of AI-generated videos.

{
    \small
    \bibliographystyle{ieeenat_fullname}
    \bibliography{main}
}

\clearpage
\setcounter{page}{1}
\maketitlesupplementary

\section*{Experiment Setting}
\label{appendix: Exp}
\subsection*{Dataset}
\label{appendix: dataset}
\paragraph{Training data} The training data in our experiment comes from OpenVid-1M~\citep{nan2024openvid}, which comprises over 1 million in-the-wild video clips, all with resolutions of at least 512×512, accompanied by detailed captions. We download a subset of it for our training. Specifically, we randomly select 10k videos from the downloaded videos for training. Each training video is composed of 8 frames sampled with a frame interval of 8. During fine-tuning, the videos are resized to \(256\times 256\) for lower computational overhead.

\paragraph{Evaluation data} As we mentioned in Section~\ref{section: Exp settings}, we select 
50 prompts from the test set of VBench~\citep{huang2024vbench}, covering five categories: Animal, Human, Plant, Scenery, and Vehicles (10 prompts each). We list these prompts in Table~\ref{tab:prompt-categories}. For each prompt, we generate four videos using fixed random seeds (42, 114514, 3407, 6666) to ensure reproducibility. For the I2V task, we first generate an image using Stable Diffusion 2.1~\citep{rombach2022high}, and subsequently generate four videos conditioned on each generated image using the same set of random seeds. Each generated video contains 16 frames with a resolution \(512\times512\).

\subsection*{Training Detail} 
\label{appendix: training detail}
We use an AdamW optimizer for fine-tuning, with a learning rate \(5\times10^{-4}\) and batch size 2. We adopt a learning rate schedule with a linear warm-up followed by cosine decay. Specifically, the learning rate increases linearly from 0 to the base learning rate \( \text{lr}_{\text{b}} \) over the first \( T_{\text{w}} \) warmup steps, and then decays to a minimum value \( \text{lr}_{\text{m}} = 10^{-6} \) following a half-cycle cosine schedule:

\begin{equation}
g(t) = 1 + \cos\left(
    \frac{\pi (t - T_{\text{w}})}{T_{\text{t}} - T_{\text{w}}}
\right)
\end{equation}

\begin{equation}
\text{lr}(t) =
\begin{cases}
\dfrac{t}{T_{\text{w}}}\,\text{lr}_{\text{b}}, & t < T_{\text{w}} \\[4pt]
\text{lr}_{\text{m}} + \dfrac{1}{2}(\text{lr}_{\text{b}} - \text{lr}_{\text{m}})\, g(t),
& t \ge T_{\text{w}}
\end{cases}
\end{equation}

where \( t \) is the current training step, \( T_{\text{t}} \) is the total number of training steps, and \( \text{lr}_{\text{b}} \) is the base learning rate \(5\times10^{-4}\). For parameter groups with an individual scaling factor \( \text{lr}_{\text{scale}} \), the learning rate is further scaled by this factor. In our experiments, the warm-up step \(T_\text{w}\) is set to 20\% of \( T_\text{t} \).

\subsection*{Selective Fine-tuning}

\begin{figure}
    \centering
    \includegraphics[width=\linewidth]{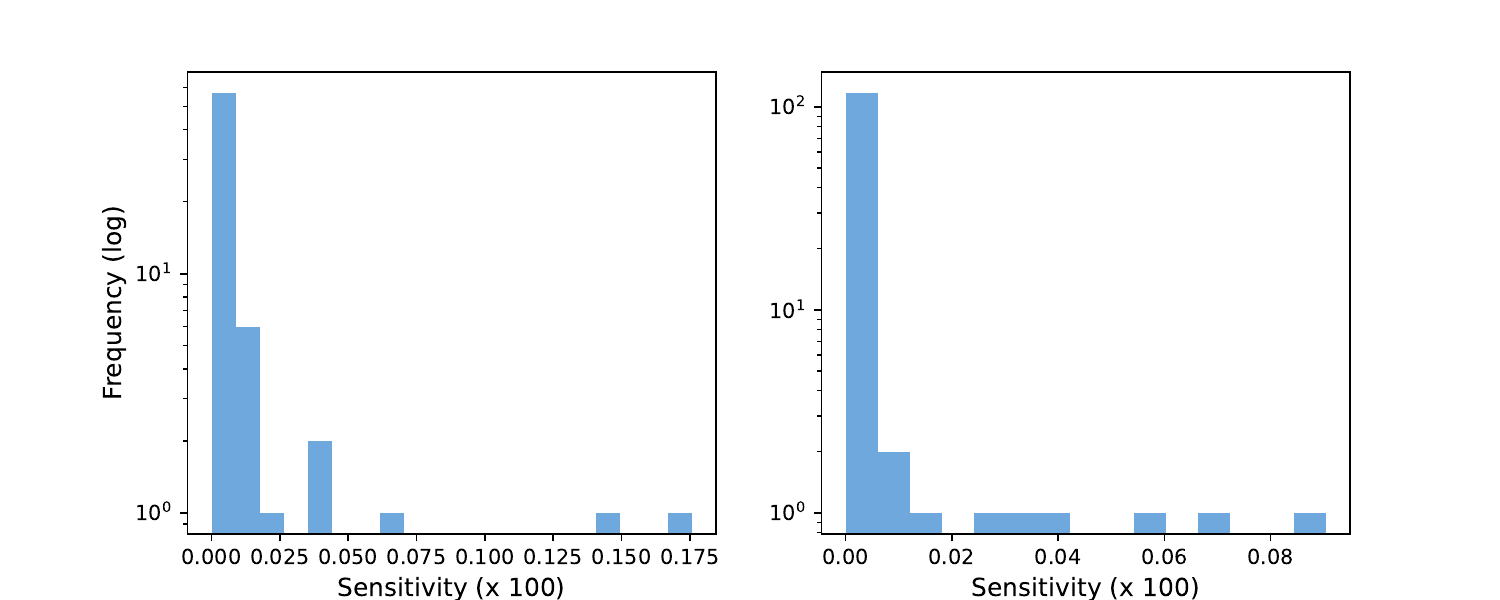}
    \caption{Histogram of layer sensitivity distributions in the latent decoder. The left figure corresponds to ModelScope (2D VAE), and the right figure corresponds to SVD (3D VAE). The perceptual sensitivity for each layer is measured by MSE.}
    \label{fig:sensitivity}
\end{figure}

Before we fine-tune the latent decoder, we apply PAS, which we proposed in this paper, to search and freeze the perceptual sensitive layers. We show the results in Figure~\ref{fig:sensitivity}. We observe that the majority of layers exhibit very low sensitivity, with values clustered near zero, indicating that perturbing these layers has minimal impact on visual quality. Only a small number of layers show relatively high sensitivity, suggesting they are more visually critical. Notably, the latent decoder of SVD demonstrates even stronger sparsity. These findings validate the design of our Perturbation-Aware Suppression (PAS) strategy: we identify and exclude perceptually sensitive layers and selectively fine-tune only the insensitive ones, enabling effective watermark embedding with minimal visual degradation. In our experiments, we set the threshold \(\tau_1 = 1.5\times10^{-4}\) for ModelScope and \(\tau_2 = 10^{-4}\) for SVD. 

\subsection*{Temporal Tampering implementation}
\label{appendix: Temporal tampering}
In this paper, to test the resilience of \textsc{VidSig}, we implement several temporal attacks. We provide detailed descriptions of these attacks here. Give a video \( [f_1, f_2, f_3, \ldots, f_{N}] \) consists of \(N\) frames, we define the following temporal attacks.

\paragraph{Frame Drop} Randomly select a frame and delete it. Let \(i\) denote the index of the selected frame, the tampered video is then denoted as \([f_1, f_2, f_{i-1}, f_{i+1}, \ldots, f_{N}] \)

\paragraph{Frame Swap} We randomly select two distinct frames, indexed by \(i\) and \(j\), and swap their positions in the video, resulting in a perturbed sequence \( [f_1, \ldots, f_{i-1}, f_j, \ldots, f_{j-1}, f_i, \ldots, f_N] \).

\paragraph{Frame Insert} We randomly select two positions indexed by \(i\) and \(j\), and insert the \(i^{\text{th}}\) frame at position \(j\), resulting in a new sequence with duplicated content and increased temporal length \( [f_1, f_2, f_i, \ldots f_{j-1}, f_i, f_j \ldots, f_{N+1}] \)

\paragraph{Frame Insert Gaussian} Randomly select a position indexed by \(i\), and insert a standard Gaussian noise at position \(i\), denoted as \(f_i'\), resulting in a new sequence with increased temporal length \( [f_1, f_2, \ldots, f_i',f_i  \ldots, f_{N+1}] \)

\paragraph{Frame Average} Given a sequence length \(n\), we randomly select a position indexed by \(i\), conditioned on \(i + n \leq N\). We then compute the average of the \(n\) subsequent frames, i.e., \( \bar{f} = \frac{1}{n} \sum_{j=0}^{n-1} f_{i+j} \), and replace the \(i^{\text{th}}\) frame with \(\bar{f}\), resulting the tampered video \( [f_1, f_2, \ldots, f_{i-1}, \bar{f},\ldots, f_{N-n+1}] \)

\section*{Metric}
\label{appendix:metric}
\subsection*{Watermark Detection}
Let \( \mathbf{m} \in \{0, 1\}^k \) be the embedded multi-bit message, and let \( \mathbf{m}' \) be the extracted message from a generated video \( \mathbf{v} \). To determine whether the video contains the watermark, we count the number of matching bits between \( \mathbf{m} \) and \( \mathbf{m}' \), denoted as \( M(\mathbf{m}, \mathbf{m}') \). A detection decision is made by checking whether the number of matching bits exceeds a predefined threshold \( \tau \):

\begin{equation}
M(\mathbf{m}, \mathbf{m}') \geq \tau, \quad \text{where } \tau \in \{0, \dots, k\}.
\end{equation}

To formally evaluate detection performance, we define the hypothesis test as follows:  
\( H_1 \): "The video was generated by the watermarked model" vs.  
\( H_0 \): "The video was not generated by the watermarked model".
Under the null hypothesis \( H_0 \), as in privious work\citep{yu2021artificial}, assume that the extracted bits \( m'_1, \dots, m'_k \) are i.i.d. Bernoulli variables with \(p=0.5\), the match count \( M(\mathbf{m}, \mathbf{m}') \) follows a binomial distribution \( \mathcal{B}(k, 0.5) \). The False Positive Rate (FPR) is defined as the probability of falsely detecting a watermark when \( H_0 \) is true:

\begin{equation}
\mathrm{FPR}(\tau) = \mathbb{P}(M > \tau \mid H_0) = I_{1/2}(\tau + 1, k - \tau),
\end{equation}

where \( I_{x}(a, b) \) is the regularized incomplete beta function. Based on this, we can calculate the least match count to determine whether the video contains watermark under a fixed FPR.

\begin{table*}[t]
\centering
\caption{Prompts for video generation across five domains.}
\label{tab:prompt-categories}
\renewcommand{\arraystretch}{1.15}
\setlength{\tabcolsep}{4pt}
\resizebox{\textwidth}{!}{
\begin{tabular}{p{2.2cm}p{5.9cm}p{5.9cm}}
\toprule
\textbf{Category} & \textbf{Prompt (1)} & \textbf{Prompt (2)} \\
\midrule
\multirow{5}{*}{\textbf{Animal}} 
& a red panda eating leaves & a squirrel eating nuts \\
& a cute pomeranian dog playing with a soccer ball & curious cat sitting and looking around \\
& wild rabbit in a green meadow & underwater footage of an octopus in a coral reef \\
& hedgehog crossing road in forest & shark swimming in the sea \\
& an african penguin walking on a beach & a tortoise covered with algae \\
\midrule
\multirow{5}{*}{\textbf{Human}} 
& a boy covering a rose flower with a dome glass & boy sitting on grass petting a dog \\
& a child playing with water & couple dancing slow dance with sun glare \\
& elderly man lifting kettlebell & young dancer practicing at home \\
& a man in a hoodie and woman with a red bandana talking to each other and smiling & a woman fighter in her cosplay costume \\
& a happy kid playing the ukulele & a person walking on a wet wooden bridge \\
\midrule
\multirow{3}{*}{\textbf{Plant}} 
& plant with blooming flowers & close up view of a white christmas tree \\
& dropping flower petals on a wooden bowl & a close up shot of gypsophila flower \\
& a stack of dried leaves burning in a forest &  drone footage of a tree on farm field \\
&shot of a palm tree swaying with the wind  & candle wax dripping on flower petals   \\
& forest trees and a medieval castle at sunset & a mossy fountain and green plants in a botanical garden \\
\midrule
\multirow{6}{*}{\textbf{Scenery}} 
& scenery of a relaxing beach & fireworks display in the sky at night \\
& waterfalls in between mountain & exotic view of a riverfront city \\
& scenic video of sunset & view of houses with bush fence under a blue and cloudy sky \\
& boat sailing in the ocean  & view of golden domed church  \\
& a blooming cherry blossom tree under a blue sky with white clouds & aerial view of a palace \\
\midrule
\multirow{6}{*}{\textbf{Vehicles}} 
& a helicopter flying under blue sky & red vehicle driving on field \\
& aerial view of a train passing by a bridge & red bus in a rainy city \\
& an airplane in the sky & helicopter landing on the street \\
& boat sailing in the middle of the ocean & video of a kayak boat in a river \\
& traffic on busy city street & slow motion footage of a racing car  \\
\bottomrule
\end{tabular}
}
\end{table*}
\subsection{Video Quality}
We provide details of the metrics for video quality evaluation in our experiments. Specifically, we utilize Peak Signal-to-Noise Ratio (PSNR), Structural Similarity Index Measure (SSIM)~\citep{wang2004image}, Learned Perceptual Image Patch Similarity (LPIPS)~\cite{zhang2018unreasonable} with a VGG backbone, tLP~\citep{chu2020learning} and the VBench evaluation metrics~\citep{huang2024vbench}. PSNR, SSIM, LPIPS, and tLP are computed between the watermarked video and its corresponding unwatermarked version to assess fidelity, whereas VBench evaluates the perceptual quality of the generated video independently.

\paragraph{PSNR}
PSNR quantifies the pixel-wise difference between the generated watermarked video \(\hat{\mathbf{v}}\) and the original video \(\mathbf{v}\). It is computed as the average PSNR across all frames:

\begin{equation}
    \mathrm{PSNR} = \frac{1}{f} \sum_{t=1}^{f} 10 \cdot \log_{10} \left( \frac{\mathrm{MAX}^2}{\mathrm{MSE}_t} \right),
\end{equation}

where \(f\) is the number of frames, \(\mathrm{MAX}\) denotes the maximum possible pixel value (typically 1.0 or 255), and \(\mathrm{MSE}_t = \frac{1}{mn} \sum_{i=1}^{m} \sum_{j=1}^{n} \left[\hat{\mathbf{v}}_t(i,j) - \mathbf{v}_t(i,j)\right]^2\) is the mean squared error between the \(t\)-th frame of the two videos. Higher PSNR values indicate better fidelity to the original video. 

\paragraph{SSIM}  
SSIM measures the structural similarity between the generated watermarked video \(\hat{\mathbf{v}}\) and the original video \(\mathbf{v}\). The overall SSIM is computed by averaging the SSIM values of all frames:

\begin{equation}
\mathrm{SSIM} = \frac{1}{f} \sum_{t=1}^{f} \mathrm{SSIM}(\hat{\mathbf{v}}_t, \mathbf{v}_t),    
\end{equation}

where \(\hat{\mathbf{v}}_t\) and \(\mathbf{v}_t\) denote the \(t\)-th frame of the watermarked and original video, respectively. The SSIM between two frames is defined as:

\begin{equation}
    \mathrm{SSIM}(x, y) = \frac{(2\mu_x\mu_y + C_1)(2\sigma_{xy} + C_2)}{(\mu_x^2 + \mu_y^2 + C_1)(\sigma_x^2 + \sigma_y^2 + C_2)},
\end{equation}

where \(\mu_x, \mu_y\) are the means, \(\sigma_x^2, \sigma_y^2\) are the variances, and \(\sigma_{xy}\) is the covariance of frame patches from \(x\) and \(y\); \(C_1, C_2\) are constants to stabilize the division. SSIM ranges from 0 to 1, with higher values indicating greater perceptual similarity.

\begin{table*}[tbp]
\centering
\caption{Specific VBench score for each watermarking method.}
\resizebox{\textwidth}{!}{
\begin{tabular}{c|lcccc}
\toprule
Model & Method & Subject Consistency & Background Consistency & Motion Smoothness &Imaging Quality \\
\midrule
\multirow{5}{*}{MS}
  & RivaGAN          &   0.955   &  0.963 & 0.980 & 0.675 \\
  & VideoSeal         & 0.955	 & 0.962  & 0.980 & 0.676 \\
  & StableSig        &0.956 & 0.961 & 0.980 & 0.675 \\
  & VideoShield     & 0.952 & 0.962   & 0.977 & 0.683  \\
  & \textsc{VidSig} (ours) &  0.956 & 0.961 & 0.978 & 0.676\\
\midrule
\multirow{5}{*}{SVD} 	 	
  & RivaGAN          & 0.945     & 0.958 & 0.956  &  0.623 \\
  & VideoSeal        & 0.945      & 0.955  &0.956 & 0.624 \\
  & StableSig        & 0.946  & 0.956  & 0.957  &0.633  \\
  & VideoShield      & 0.934 & 0.949 & 0.956 & 0.629  \\
  & \textsc{VidSig} (ours) & 0.946 &0.955	&0.959&	0.632\\ 
\bottomrule
\end{tabular}
}
\label{app tab: video quality}
\end{table*}

\paragraph{LPIPS}
LPIPS~\citep{zhang2018unreasonable} evaluates perceptual similarity by computing deep feature distances between frames. Given a ground-truth video \(\mathbf{v} = \{ \mathbf{v}_1, \ldots, \mathbf{v}_f \}\) and a generated video \(\hat{\mathbf{v}} = \{ \hat{\mathbf{v}}_1, \ldots, \hat{\mathbf{v}}_f \}\), the video-level LPIPS score is defined as the average over all frames:

\begin{equation}
\mathrm{LPIPS} = \frac{1}{f} \sum_{t=1}^{f} \mathrm{LPIPS}(\hat{\mathbf{v}}_t, \mathbf{v}_t),    
\end{equation}

where \(\mathrm{LPIPS}(\hat{\mathbf{v}}_t, \mathbf{v}_t)\) denotes the perceptual distance between the \(t\)-th frame pair, computed via a pretrained deep network. Lower values indicate higher perceptual similarity.

\paragraph{tLP}
tLP~\citep{chu2020learning} measures the consistency of temporal perceptual dynamics between adjacent frames. Specifically, it compares the learned perceptual (LP) differences between adjacent frame pairs in the generated video \(\hat{\mathbf{v}} = \{ \hat{\mathbf{v}}_1, \ldots, \hat{\mathbf{v}}_f \}\) and its reference \(\mathbf{v} = \{ \mathbf{v}_1, \ldots, \mathbf{v}_f \}\). The tLP is defined as:

\begin{equation}
\mathrm{tLP} = \left\| \mathrm{LP}(\hat{\mathbf{v}}_{t-1}, \hat{\mathbf{v}}_t) - \mathrm{LP}(\mathbf{v}_{t-1}, \mathbf{v}_t) \right\|_1,
\end{equation}

where \(\mathrm{LP}(\cdot, \cdot)\) denotes the LPIPS distance between two frames. A lower tLP indicates better temporal coherence with respect to the ground-truth dynamics.

\paragraph{VBench: Subject Consistency} 
Subject Consistency evaluates the semantic stability of the generated subject by calculating the cosine similarity of DINO~\citep{caron2021emerging} features across frames:

\begin{equation}
S_{\text{subject}} = \frac{1}{T-1} \sum_{t=2}^{T} \frac{1}{2} \left( \langle d_1, d_t \rangle + \langle d_{t-1}, d_t \rangle \right),
\end{equation}

where \( d_i \) is the normalized DINO image feature of the \( i^{th} \) frame, and \( \langle \cdot, \cdot \rangle \) denotes cosine similarity via dot product.

\paragraph{VBench: Background Consistency}
Background Consistency measures the temporal consistency of the background using CLIP~\citep{radford2021learning} features:

\begin{equation}
S_{\text{background}} = \frac{1}{T-1} \sum_{t=2}^{T} \frac{1}{2} \left( \langle c_1, c_t \rangle + \langle c_{t-1}, c_t \rangle \right),
\end{equation}

where \( c_i \) is the CLIP image feature of the \(i^{th}\) frame, normalized to unit length.

\paragraph{VBench: Motion Smoothness}
Given a video \( [f_0, f_1, f_2, \ldots, f_{2n}] \), the odd-numbered frames \( [f_1, f_3, \ldots, f_{2n-1}] \) are dropped, and the remaining even-numbered frames \( [f_0, f_2, \ldots, f_{2n}] \) are used to interpolate~\citep{li2023amt} intermediate frames \( [\hat{f}_1, \hat{f}_3, \ldots, \hat{f}_{2n-1}] \). The Mean Absolute Error (MAE) between interpolated and original dropped frames is computed and normalized to \([0, 1]\), with a larger value indicating better smoothness.

\paragraph{VBench: Imaging Quality}
Imaging quality mainly considers the low-level distortions presented in the generated video
frames. The MUSIQ~\citep{ke2021musiq} image quality predictor trained on the SPAQ~\citep{fang2020perceptual} dataset is used for evaluation, which is capable of handling variable sized aspect ratios and resolutions. The frame-wise score is linearly normalized to [0, 1], and the final score is then calculated by averaging the frame-wise scores across the entire video sequence.

\begin{figure}
    \centering
    \includegraphics[width=\linewidth]{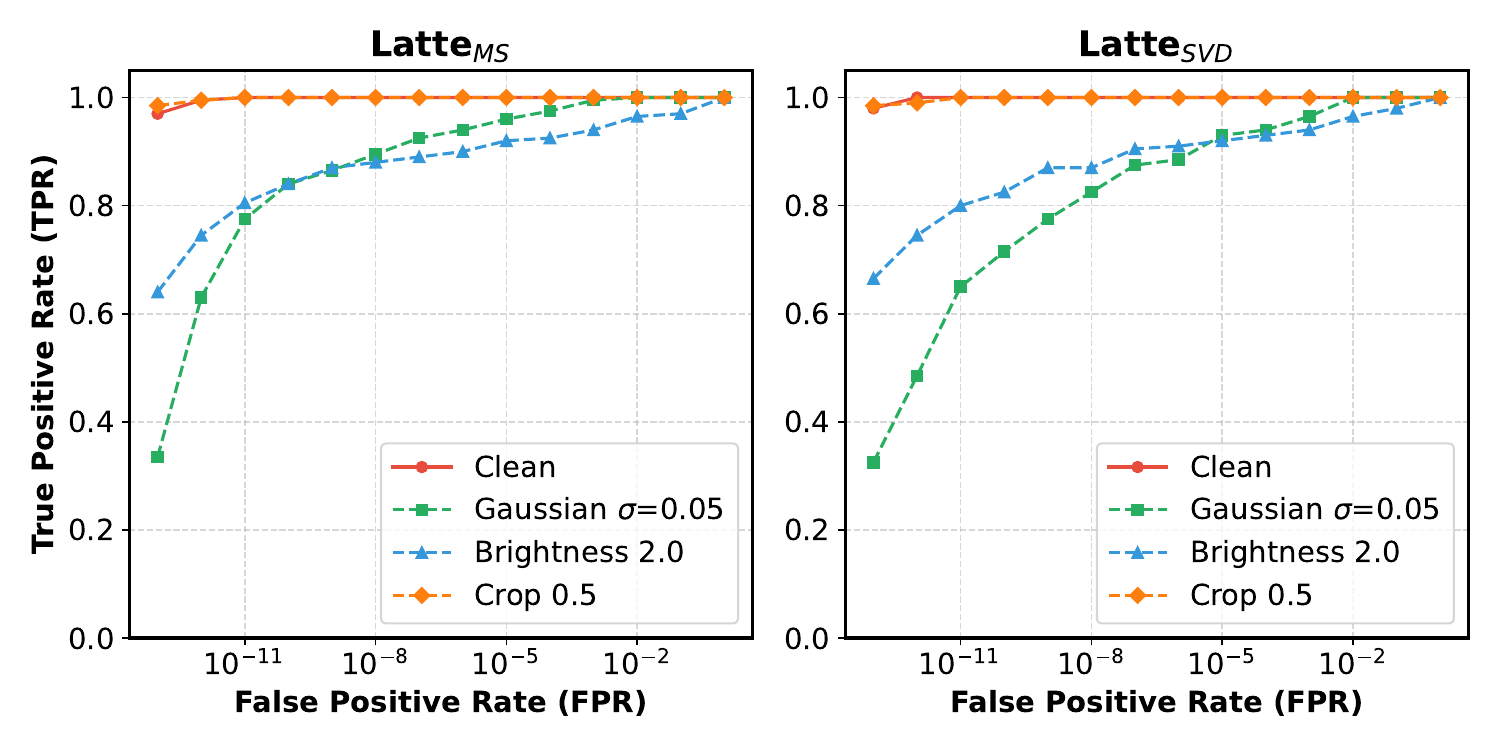}
    \caption{Watermark detection Results for \(\text{Latte}_{MS}\) and  \(\text{Latte}_{SVD}\).}
    \label{fig:latte_tpr}
\end{figure}

\section*{More Experiment Results}

\subsection*{Video Quality}
We provide the specific video quality metric of VBench for each watermarking method in Table~\ref{app tab: video quality}, It can be observed that \textsc{VidSig} achieves comparable or even superior video quality compared to post-generation methods, while significantly outperforming them in watermark extraction and detection accuracy, as shown in Table~\ref{tab:main results}.

\subsection*{Watermark Detection}

In Section~\ref{sec: EXP transfer}, we substitute the latent decoder of Latte~\citep{ma2024latte} by the fine-tuned latent decoders in MS and SVD. We also follow the same settings in Section~\ref{section: main results} and report the True Positive Rate, the results are shown in Figure~\ref{fig:latte_tpr}. 

\subsection*{Detailed Video Quality and TPR under Different Generation Configurations}
\label{appendix:config-table}

Table~\ref{tab:quality_avg1} and Table~\ref{tab:quality_avg2} reports the detailed average video quality (Avg) scores and TPR@1e-4 for both the T2V model (ModelScope) and the I2V model (SVD) across different frame numbers and resolutions.

\begin{table}[t]
\centering
\small
\caption{Detailed Avg video quality under varying frame numbers.}
\begin{tabular}{c|c|ccc}
\toprule
Model & Type & Frame 16 & Frame 24 & Frame 32 \\
\midrule
\multirow{3}{*}{MS}
  & Vanilla     & 0.897 & 0.840 & 0.832 \\
  & Watermarked  & 0.893 & 0.842 & 0.835 \\
  & TPR@1e-4  & 1.000 & 0.995 & 0.965 \\
\midrule
\multirow{3}{*}{SVD}
  & Vanilla     & 0.872 & 0.864 & 0.855 \\
  & Watermarked  & 0.873 & 0.865 & 0.855 \\
  & TPR@1e-4  & 1.000 & 1.000 & 1.000 \\
\bottomrule
\end{tabular}
\label{tab:quality_avg1}
\end{table}

\begin{table}[t]
\centering
\small
\caption{Detailed Avg video quality under varying resolutions.}
\resizebox{\linewidth}{!}{
\begin{tabular}{c|c|cccc}
\toprule
Model & Type & Resolution 256 & Resolution 512 & Resolution 768 &Resolution 1024 \\
\midrule
\multirow{3}{*}{MS}
  & Vanilla     & 0.871 & 0.897 & 0.892 & 0.900 \\
  & Watermarked  & 0.870 & 0.893 & 0.892 & 0.901\\
  & TPR@1e-4  & 1.000 & 1.000 & 1.000 & 0.990\\
\midrule
\multirow{3}{*}{SVD}
  & Vanilla     & 0.719 & 0.872 & 0.886 & 0.887\\
  & Watermarked  & 0.719 & 0.873 & 0.886 & 0.888\\
  & TPR@1e-4  & 1.000 & 1.000 & 1.000 & 1.000\\
\bottomrule
\end{tabular}
}
\label{tab:quality_avg2}
\end{table}

\begin{figure}[t]
    \centering
    \includegraphics[width=0.8\linewidth]{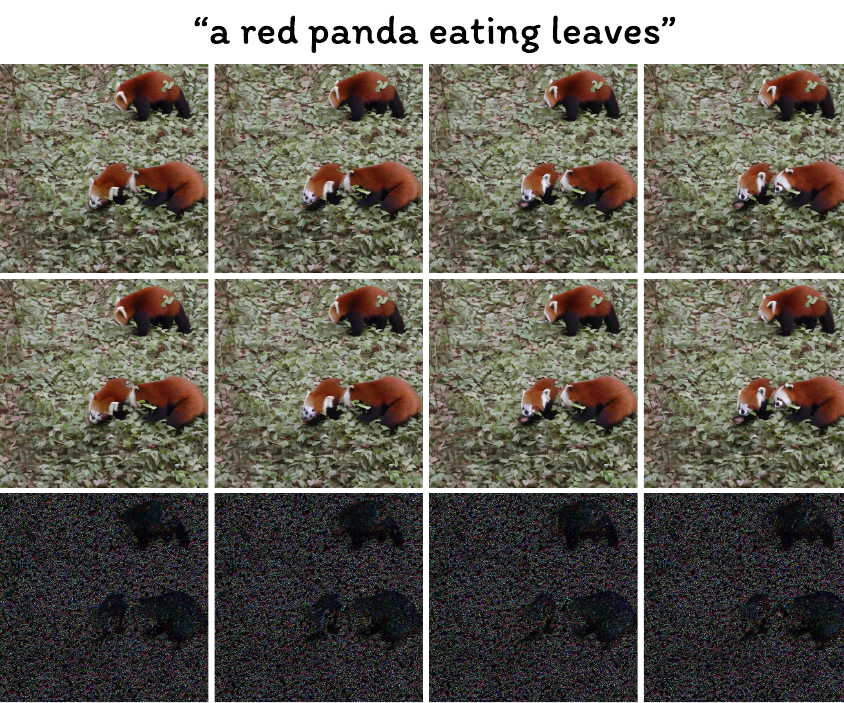}
    \caption{Qualitative comparison of the original and watermarked videos. The first row is the original video, the second row is the video generated by \textsc{VidSig}, and the third row refers to the pixel-wise difference (\(\times 10\)). 
    }
    \label{fig:case study}
\end{figure}

\subsection*{Qualitative Analysis} The qualitative comparison between the original and watermarked videos is shown in Figure~\ref{fig:case study}. It's clear to see that the watermark manifests only as imperceptible high‑frequency perturbations, with no structured artefacts or semantic drift observable across time. It's also noticed that in real-world applications, all of the generated videos will have automatically embedded watermarks, thus there will be no reference videos for comparison. The VBench scores in Table~\ref{tab:main results} reveal that \textsc{VidSig} will not degrade the video quality, even though metrics like PSNR and SSIM are lower than post-generation methods.

\subsection*{More Visual Cases}
\begin{figure}
    \centering
    \includegraphics[width=\linewidth]{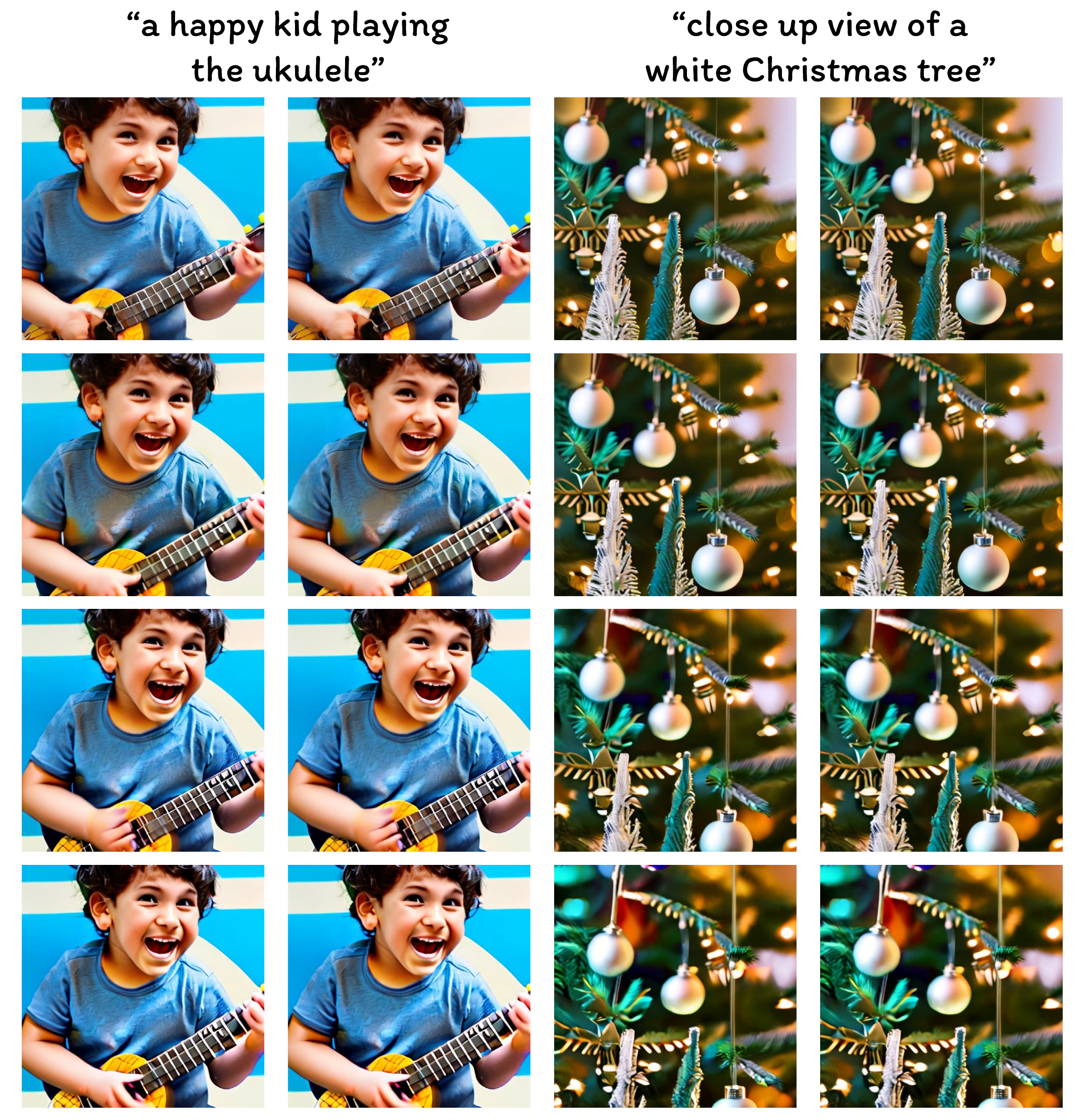}
    \caption{Two videos generated by Stable Video Diffusion, the left one is the original video, and the right one is the corresponding watermarked video by \textsc{VidSig}.}
    \label{fig:video case svd}
\end{figure}

\begin{figure}
    \centering
    \includegraphics[width=\linewidth]{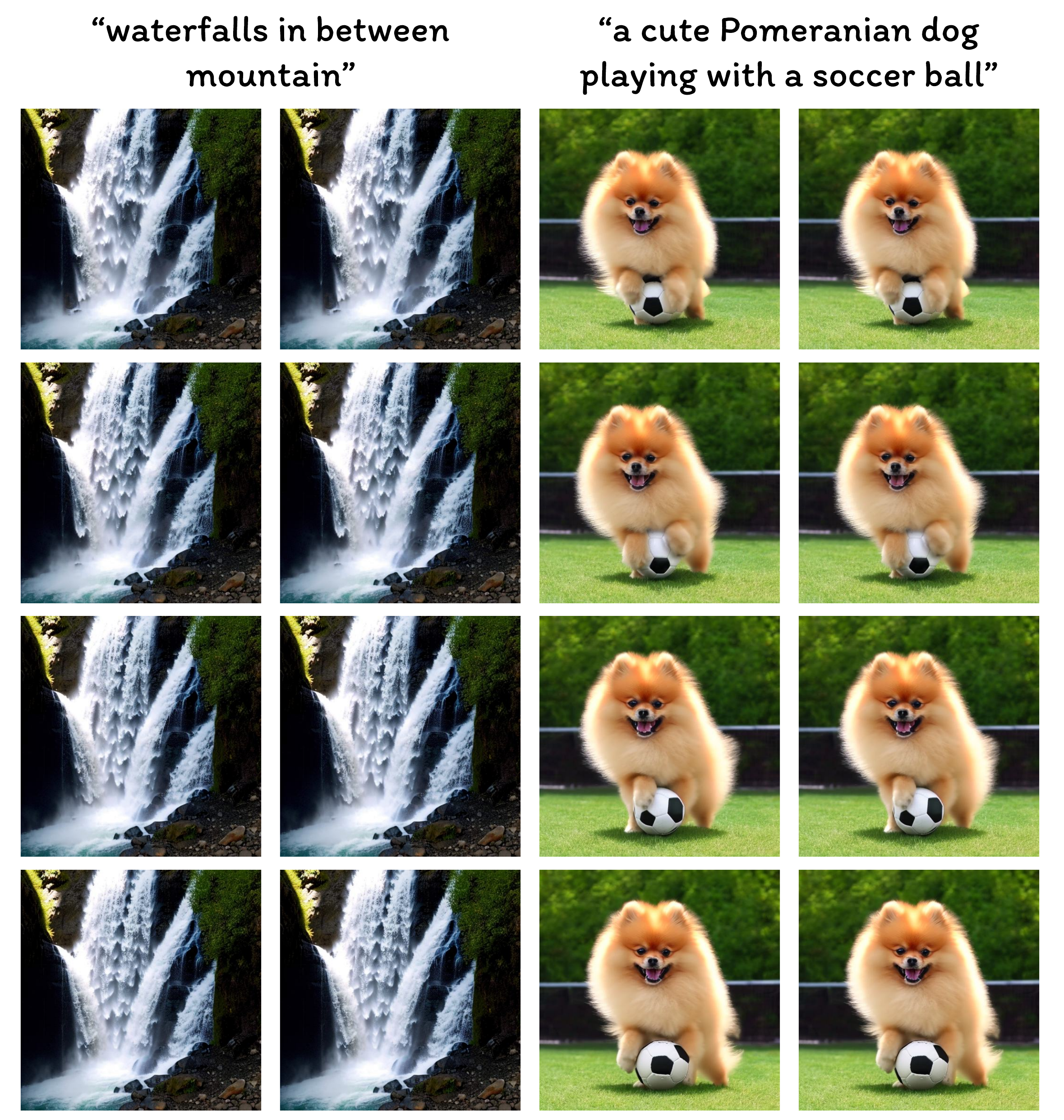}
    \caption{Two videos generated by Latte, the left one is the original video, and the right one is the corresponding watermarked video by \textsc{VidSig}. The latent decoder of Latte is transferred from ModelScope.}
    \label{fig:video case lattems}
\end{figure}

\begin{figure}
    \centering
    \includegraphics[width=\linewidth]{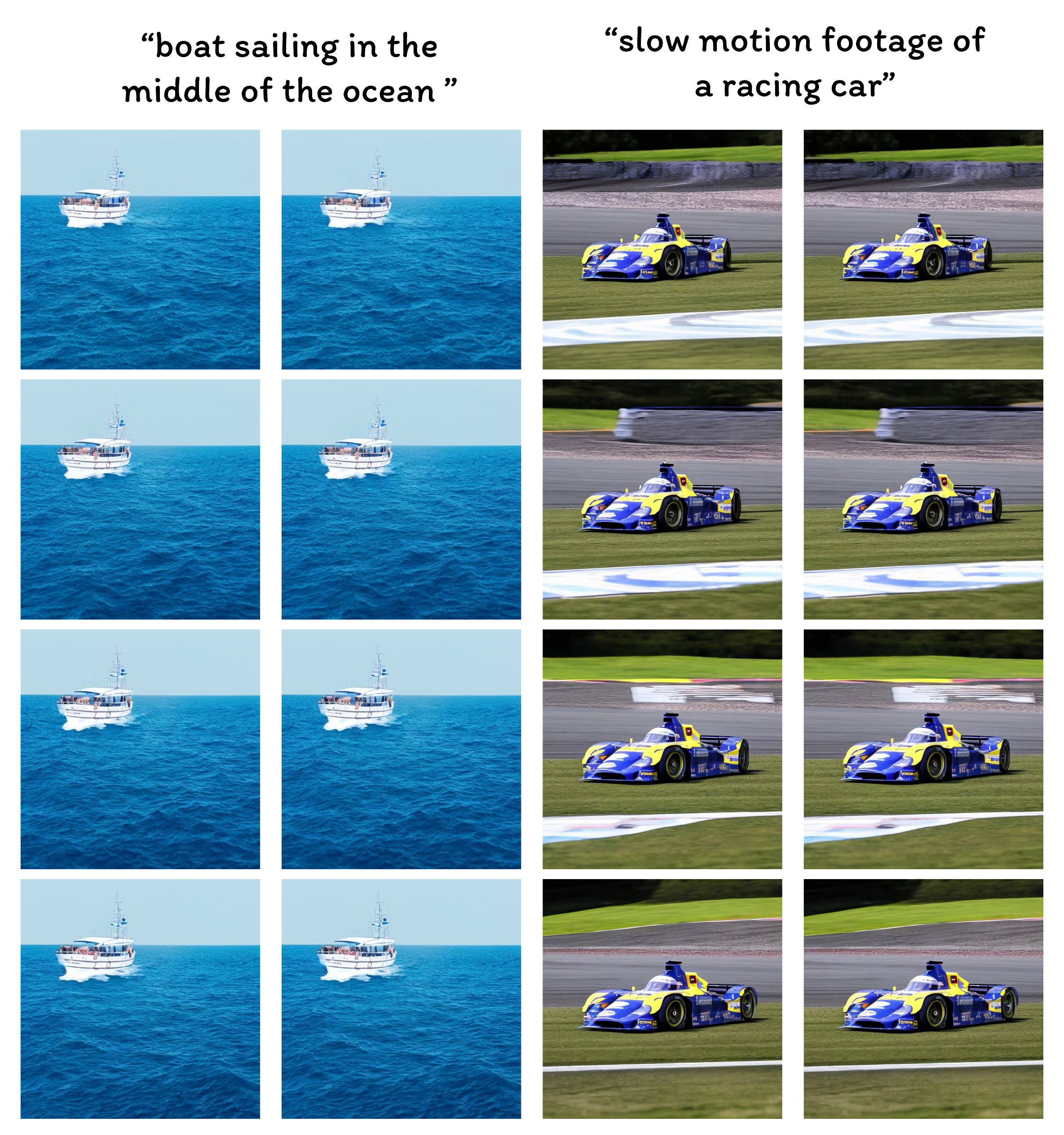}
    \caption{Two videos generated by Latte, the left one is the original video, and the right one is the corresponding watermarked video by \textsc{VidSig}. The latent decoder of Latte is transferred from Stable Video Diffusion.}
    \label{fig:video case lattesvd}
\end{figure}

We show more videos generated by SVD and Latte here, see Figure~\ref{fig:video case svd}, Figure~\ref{fig:video case lattems}, and Figure~\ref{fig:video case lattesvd}.

% \section{Rationale}
% \label{sec:rationale}
% % 
% Having the supplementary compiled together with the main paper means that:
% % 
% \begin{itemize}
% \item The supplementary can back-reference sections of the main paper, for example, we can refer to \cref{sec:intro};
% \item The main paper can forward reference sub-sections within the supplementary explicitly (e.g. referring to a particular experiment); 
% \item When submitted to arXiv, the supplementary will already included at the end of the paper.
% \end{itemize}
% % 
% To split the supplementary pages from the main paper, you can use \href{https://support.apple.com/en-ca/guide/preview/prvw11793/mac#:~:text=Delete%20a%20page%20from%20a,or%20choose%20Edit%20%3E%20Delete).}{Preview (on macOS)}, \href{https://www.adobe.com/acrobat/how-to/delete-pages-from-pdf.html#:~:text=Choose%20%E2%80%9CTools%E2%80%9D%20%3E%20%E2%80%9COrganize,or%20pages%20from%20the%20file.}{Adobe Acrobat} (on all OSs), as well as \href{https://superuser.com/questions/517986/is-it-possible-to-delete-some-pages-of-a-pdf-document}{command line tools}.

\end{document}